%% file: main.tex
\PassOptionsToPackage{table,dvipsnames}{xcolor}
\RequirePackage{amsmath}
\documentclass{applemlr}

\DeclareFontShape{T1}{sfpro}{m}{sc}{<-> s*[0.88] store/SFPro-Regular}{}
\DeclareFontShape{T1}{sfpro}{b}{sc}{<-> s*[0.88] store/SFPro-Bold}{}
\DeclareFontShape{T1}{sfpro}{bx}{sc}{<-> s*[0.88] store/SFPro-Bold}{}

\input{math_commands.tex}

\input{figures/figure_commands.tex}

\usepackage{hyperref}
\usepackage{url}
\usepackage{booktabs}
\usepackage{wrapfig}
\usepackage{xspace}

\newcommand{\name}{\textsc{FastFrames}\xspace}
\newcommand{\targetlatency}{50 ms}

\title{\name: Video Generation at\\Interactive Latency}

\author[1,2]{Gene Chou}
\author[1]{Guandao Yang}
\author[1]{Fujun Luan}
\author[1]{Lars Mescheder}
\author[1]{Wei Dong}
\author[1]{Stephan R. Richter}
\author[1]{Vladlen~Koltun}
\affiliation[1]{Apple}
\affiliation[2]{Cornell University}
\contribution[*]{This work was completed during Gene's internship at Apple.}
\abstract{\input{sections/abstract_apple}}

\usepackage[table]{xcolor}
\definecolor{latencyFast}{HTML}{DDEFE7}
\definecolor{latencySubsecond}{HTML}{FFF0CC}
\usepackage{threeparttable}

\newcommand{\latfast}[1]{\cellcolor{latencyFast}#1}
\newcommand{\latsub}[1]{\cellcolor{latencySubsecond}#1}

\begin{document}
\maketitle

\input{sections/01_introduction}

\input{figures/fig_pipeline}
\input{sections/02_related}

\input{sections/03_method}
\input{sections/04_results}

\input{sections/05_conclusion}

\FloatBarrier
\section*{Acknowledgments}
We thank Hongyu Wen, Boyang Deng, Alejandro Newell, Aleksei Bochkovskii, JunYoung Gwak, Lahav Lipson, Marcel Santos, Marko Mihajlovic, Peiyun Hu, Lyndon Duong, Amael Delaunoy, Jackson Hamburger, and Philipp Kraehenbuehl for discussions and support.

\bibliography{references}
\bibliographystyle{references}

\clearpage
\input{sections/x_supp}

\end{document}

%% file: math_commands.tex
\usepackage{amsmath,amsfonts,bm}

\def\eqref#1{equation~\ref{#1}}

\def\1{\bm{1}}

\def\vz{{\bm{z}}}

\DeclareMathAlphabet{\mathsfit}{\encodingdefault}{\sfdefault}{m}{sl}
\SetMathAlphabet{\mathsfit}{bold}{\encodingdefault}{\sfdefault}{bx}{n}

\providecommand*{\eg}{e.g.,~}

%% file: figures/figure_commands.tex
\usepackage{graphicx}
\newlength{\phgap}
\newlength{\phlblw}
\newlength{\phw}
\newlength{\phh}
\AtBeginDocument{%
  \setlength{\phlblw}{0.11\textwidth}%
  \setlength{\phw}{\dimexpr(\textwidth-\phlblw-5\phgap-1pt)/6\relax}%
  \setlength{\phh}{0.638\phw}}

\usepackage{tikz}
\definecolor{prefours}{rgb}{0.27,0.47,0.71}
\definecolor{preftie}{rgb}{0.85,0.85,0.85}
\definecolor{prefbase}{rgb}{0.60,0.62,0.65}

\newlength{\abw}
\newlength{\abh}

\newlength{\abgap}

%% file: sections/abstract_apple.tex
Autoregressive video diffusion models generate frames based on new observations and inputs, making them a promising medium for interactive visual experiences. However, existing models can be orders of magnitude too slow for fluid interaction. This paper introduces \name, a family of fast autoregressive video diffusion models that achieves interactive latency without compromising quality.
We profile the inference pipeline of state-of-the-art models and find that operations often treated as fixed implementation details, such as VAE decoding and KV cache updates, in fact account for over half of end-to-end inference latency. We challenge the assumption that these components are indispensable and replace them with lightweight alternatives. 
Our 1.3B variant runs at 36~ms per frame, or 28 responses per second (RPS), on a single B200 GPU while matching the generation quality of a state-of-the-art model that is nearly 8× slower. Our 14B variant runs at 45~ms per frame, or 22 RPS, on an 8× B200 GPU node and yields significantly better visual quality and realism. 
We show that \name enables real-time interaction in applications such as camera-conditioned and physics-conditioned video generation. See our project page for videos, code, and model weights.

\begin{center}
\vspace{0.5em}
\url{https://apple-aiml-research.github.io/ml-fastframes/}
\vspace{-3em}
\end{center}

%% file: sections/01_introduction.tex
\section{Introduction}

Video diffusion models~\citep{ho2022videodiffusion, ho2022imagenvideo, singer2022makeavideo, blattmann2023SVD} synthesize realistic videos~\citep{wan2025wan, seedance2026seedance2, blackforestlabs2026flux3, minimax2026h3}, compose diverse scenes and visual concepts~\citep{deng2025magref, chou2024kfcw, pan2026omniweaving}, and respond to camera and action controls~\citep{wang2024motionctrl, he2025cameractrl}. In particular, autoregressive (AR) models~\citep{jin2024pyramidal, deng2024nova, chen2025skyreelsv2infinitelengthfilmgenerative, ai2025magi1autoregressivevideogeneration} generate videos incrementally, allowing them to incorporate new controls or observations. 
They are a promising medium for interactive visual applications, such as world simulation~\citep{chou2026cityrag, seo2026grounding}, games~\citep{yu2025gamefactory, lingbot-world}, and predictive models for robotics~\citep{ye2026worldactionmodelszeroshot, nvidia2026cosmos3}. 

Recent works have significantly accelerated AR video diffusion models, primarily through distillation and few-step sampling~\citep{yin2024dmd2, yin2024causvid}. Some now report real-time generation speeds based on frames per second (FPS)~\citep{huang2026selfforcing, zhu2026causal}.
However, FPS alone does not establish whether a model can respond to user inputs in real time. For instance, a model that generates a 30-frame clip as a single batch in one second achieves 30 FPS, but the user must still wait one second to see a response.
Seamless real-time interactivity demands low input-to-response latency, as established in interactive systems research~\citep{miller1968response}. Human-computer interaction studies have reported delay-detection thresholds of approximately 50 to 60 ms~\citep{annett2014how, deber2015how, forch2017are}.

\newpage
\input{figures/fig_pareto}

Thus, we target \emph{end-to-end per-frame latency} below \targetlatency. We define this latency as the delay between the arrival of a new input signal (\eg a steering command) and the presentation of the first RGB frame that depicts its effects. In the baseline methods we study~\citep{yin2024causvid, huang2026selfforcing, zhao2026causal}, the inference pipeline consists of KV cache updates, diffusion sampling, and pixel rendering. Measuring latency requires accounting for the complete critical path through these operations. 

We also report the corresponding \emph{responses per second} (RPS), defined as $\mathrm{RPS}=1/L_{\mathrm{s}}=1{,}000/L_{\mathrm{ms}}$, where $L_{\mathrm{s}}$ and $L_{\mathrm{ms}}$ denote this latency in seconds and milliseconds, respectively. Each generation unit counts as one response, regardless of how many frames it produces. Unlike FPS, which amortizes generation time over the frames in a batch, RPS directly measures responsiveness for interactive applications; an interactive latency target of \targetlatency\ corresponds to 20 RPS. As shown in Fig.~\ref{fig:pareto}, the fastest baselines remain well below the target response rate.

To close this gap, we introduce \name, a family of autoregressive video diffusion models that, to the best of our knowledge, is the first to take end-to-end per-frame latency into the interactive regime ($\leq\!50$\,ms).
We analyze the limiting factors of existing models by profiling the critical inference path of a reference 1.3B-parameter model, a common model size that balances speed and generation quality
(Fig.~\ref{fig:pipeline}). We find that dominant costs lie not in diffusion denoising steps, but rather in operations often treated as fixed implementation details, such as VAE decoding and KV cache updates. For instance, a single VAE decoding step takes as long as 3.5 DiT passes.

We challenge the assumption that these components are indispensable and replace them with lightweight alternatives. 
First, we replace the VAE decoder with a ConvNeXt-based~\citep{liu2022convnet} head that operates only in low spatial resolution, followed by a single linear layer that unpatchifies and upsamples the latents to pixel dimensions, reducing decoding latency by 147× while matching quality.
Furthermore, because the head is lightweight, training the model end-to-end in pixel-space (e.g., perceptual~\citep{zhang2018perceptual} and GAN~\citep{goodfellow2014generative} losses) becomes practical, bypassing the reconstruction bottleneck of the VAE.
Second, the baseline inference pipeline includes an additional DiT evaluation on each fully denoised latent solely to refine the KV cache (Sec.~\ref{sec:kvcache}); we replace the DiT evaluation with a lightweight network, reducing cache-update latency 10x without measurable degradation in quality.
Finally, we perform multiple system-level, hardware-aware optimizations that specifically target the end-to-end per-frame critical path. 
Our techniques target components shared by many AR video diffusion models and are therefore complementary to advances in backbone architecture, such as hybrid transformers~\citep{chen2025sanavideo}, and improved distillation methods~\citep{liu2025decoupled}.

In the standard text-to-video (T2V) setting, where the model simply treats the current generated frame as input for generating the next, our quantitative and qualitative evaluations show that \name establishes a new Pareto frontier between generation quality and speed (Fig.~\ref{fig:pareto}).
The 1.3B variant achieves an end-to-end per-frame latency of 36~ms (28 RPS) on a single B200 GPU at 832×480 resolution (480p), matching the visual quality of a state-of-the-art baseline~\citep{zhao2026causal} that is 7.6× slower. Scaling \name to 14B improves visual fidelity while achieving 45~ms (22 RPS) on an 8× B200 GPU node.

The T2V AR pipeline can be extended to accept controls between generation steps, such as camera poses~\citep{jiang2026abotworld0}, 2D tracks~\citep{shin2026motionstream}, and 3D dynamics~\citep{li2025wonderplay}. 
We show that \name enables fluid interactive experiences in two applications: camera-conditioned and physics-conditioned video generation.

%% file: figures/fig_pareto.tex
\begin{wrapfigure}{r}{0.48\textwidth}
    
    \centering
    \includegraphics[width=\linewidth]{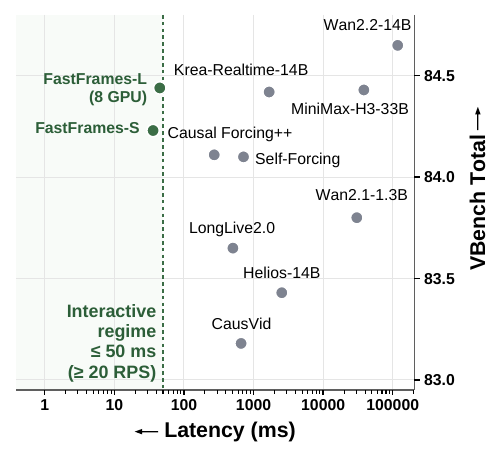}
     
    \caption{
    \textbf{Quality-speed tradeoff.}
    VBench scores vs.\ per-frame latency. Upper left is better. Only \name meets an acceptable threshold~\citep{deber2015how} for seamless interactivity
    ($\leq 50$\,ms latency, equivalently $\geq 20$ responses per second (RPS)) while matching or exceeding comparable baselines in visual quality.
    }
    \label{fig:pareto}
    \vspace{-1em}
\end{wrapfigure}

%% file: figures/fig_pipeline.tex
\begin{figure}[t]
\centering
\includegraphics[width=0.9\textwidth]{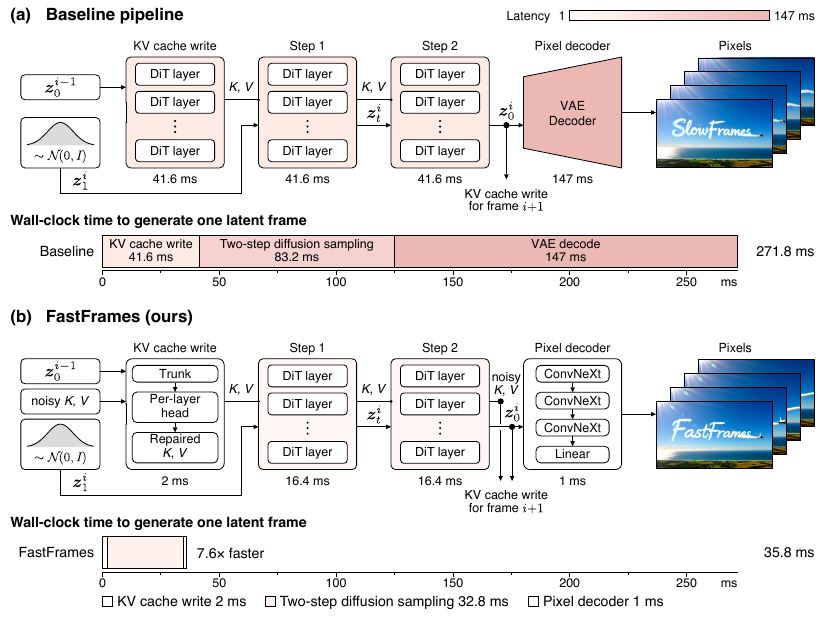}

\caption{\textbf{Inference pipeline and latency profiling for one autoregressive generation.} 
State-of-the-art few-step autoregressive video diffusion models generate each new frame by first updating the KV cache using the previous frame’s denoised latent ($\vz^{i-1}_0$) through a DiT evaluation, then performing multiple denoising steps followed by VAE decoding. The color gradient of the model components denotes speed: from white (fastest operations) to red (slowest).
\textbf{(a)}~The fastest baseline, Causal Forcing++~\citep{zhao2026causal}, takes 271 ms to generate a frame at 480p but only 30\% of wall time is spent on diffusion sampling. \textbf{(b)}~\name reduces latency nearly 8× by replacing the non-denoising ``accessories'' with lightweight alternatives and performing systems-level optimizations to speed up each DiT evaluation.
}

\label{fig:pipeline}
\end{figure}

%% file: sections/02_related.tex
\section{Related Work}

\textbf{Accelerating diffusion sampling.}
Diffusion models require iterative sampling~\citep{ho2020denoising}. Earlier work reduces the number of sampling steps using training-free accelerated samplers~\citep{song2020ddim,lu2022dpm,lu2025dpm} or by training models to take larger steps along their sampling trajectories~\citep{frans2025shortcut,geng2026meanflow,shaul2026parallel}.
The state of the art relies on distribution matching~\citep{yin2024dmd,yin2024dmd2,cui2025selfforcingpp}, where a few-step student generator approximates the distribution of a teacher network. 
Another line of work focuses on the inference cost of DiTs, including hybrid attention~\citep{li2026attend, chen2025sanavideo}, sliding-window attention for long context~\citep{xiao2024streaming, zhang2025sliding}, and layer pruning or model-size distillation~\citep{fang2024tinyfusion, wu2025individual}. A smaller line of work focuses specifically on compressing or sparsifying the KV cache of autoregressive video diffusion models to reduce attention overhead~\citep{ji2026forcingkv, liu2026diagonal, guo2026dummy}. 
However, these methods are generally model-specific, so instead we target other components on the critical path. 
Our work builds on state-of-the-art few-step models, but we observe that the iterative denoising process is no longer the latency bottleneck.

\textbf{Reducing VAE decoding time.}
State-of-the-art video diffusion models perform diffusion in latent space~\citep{blattmann2023SVD}, generally encoded via a variational autoencoder (VAE)~\citep{kingma2014autoencoding}. For instance, the Wan VAE~\citep{wan2025wan} can perform up to a 48× reduction in dimensionality. 
This makes generation cheaper but pixel decoding becomes expensive. VAE decoders repeatedly upsample features in space and time, and apply dense 3D convolutions at each resolution. For example, a tensor containing four 480×832 frames requires convolution at approximately 1.6 million spatiotemporal positions. 
A line of work aims to speed up VAEs~\citep{bohan2023taesd, buzovkin2024fastdecoders} by reducing the number of parameters and removing expensive operations like attention and wide 3D convolutions. These methods gain up to 4× speedup but incur degradation in visual quality~\citep{BoerBohan2025TAEHV, huang2026selfforcing}. Instead, we treat the pixel decoder as a lightweight readout specific to a frozen DiT, rather than a general-purpose reconstruction problem. Our pixel head maintains visual quality while being two orders of magnitude faster. 

\textbf{Systems-level optimizations.}
Systems-level techniques accelerate diffusion inference through more efficient scheduling and hardware utilization. \citet{kodaira2023streamdiffusion} and \citet{bandyopadhyay2024blockcascading} use batching and pipeline parallelism to improve streaming throughput but do not improve per-frame latency. 
StreamDiffusionV2~\citep{feng2025streamdiffusionv2} emphasizes non-amortized time-to-first-frame but to the best of our understanding, this metric focuses on the start of a stream, in contrast to our steady-state measurements.
Optimized attention kernels~\citep{dao2023flashattention2}, compilation and operator fusion~\citep{ansel2024pytorch2}, and CUDA graph execution~\citep{ghosh2025pygraph} reduce memory traffic and execution overhead. Weight and activation quantization~\citep{zhao2024viditq, li2024svdquant, su2026sixbitdiffusion, zhao2026sana} reduce memory requirements and can accelerate computation with suitable hardware support. Distributed inference spreads computation across multiple GPUs through patch, sequence, and pipeline parallelism~\citep{li2024distrifusion, fang2024pipefusion, fang2024xdit}.
Our work incorporates many of these techniques.

%% file: sections/03_method.tex
\section{Method}
\label{sec:method}
\subsection{Preliminaries: Few-Step Autoregressive Video Diffusion Models}

Autoregressive (AR) video diffusion models generate a video one latent chunk at a time, using diffusion to synthesize each chunk conditioned on previously generated ones. 
We represent a video sequence as $\vz^{1:N}=(\vz^1,\vz^2,\ldots,\vz^N)$, where each chunk $\vz^i\in\mathbb{R}^{C\times T\times H\times W}$ contains $T\geq 1$ latent frames with $C$ channels and spatial resolution $H\times W$. Each latent frame corresponds to $K$ RGB frames, where $K$ is the temporal compression factor of the VAE.
Generating larger chunks (e.g., one chunk $=$ three latent frames, as in Self-Forcing~\citep{huang2026selfforcing}) is closer to bidirectional models and empirically easier to optimize, while generating a smaller chunk (e.g., one chunk $=$ one latent frame, as in Causal Forcing++ (CF++)~\citep{zhao2026causal} and \name) is faster. 
The joint distribution factorizes autoregressively as
$p(\vz^{1:N})=\prod_{i=1}^N p(\vz^i \mid \vz^{<i}),$
where $\vz^{<i}=(\vz^1,\ldots,\vz^{i-1})$ denotes the preceding chunks.
Each conditional distribution is modeled by a diffusion transformer (DiT), which generates the next chunk by iteratively denoising an initial noise sample and attends to the preceding chunks via a KV cache.
State-of-the-art models first convert a pretrained bidirectional teacher model into an AR model, and then distill it into a few-step AR student using distribution-matching distillation~\citep{yin2024causvid, huang2026selfforcing, zhu2026causal}. We refer readers to these works for training details.

To examine the inference cost, we profile CF++, the fastest few-step AR video diffusion model at the time of writing. It is built on the Wan 2.1 1.3B-parameter backbone. Fig.~\ref{fig:pipeline} shows its inference latency on a single B200 GPU, decomposed into three stages: KV-cache updates, diffusion sampling, and pixel rendering.
We use its two-diffusion-step variant, which provides the best speed-quality tradeoff; reducing sampling steps to one significantly degrades visual quality while merely reducing latency by 15\%.
Each DiT evaluation takes 41.6 ms using BF16 precision, cuDNN attention backend~\citep{nvidia2026cudnnattention}, and \texttt{torch.compile}~\citep{ansel2024pytorch2}. The KV cache update is an additional DiT evaluation. Finally, the VAE decoder takes 147 ms to decode each generated latent into four RGB frames, resulting in a total end-to-end latency of 271.8 ms.

This profile suggests that our latency budget of \targetlatency\ leaves room for only a single diffusion sampling step. 
Instead, we replace the VAE with lightweight pixel rendering (Sec.~\ref{sec:decoder}) and accelerate KV cache updates (Sec.~\ref{sec:kvcache}). Then, by reducing the latency of each DiT evaluation through systems-level optimizations (Sec.~\ref{sec:sysopt}), we fit the full inference pipeline within the target latency.
Our final 1.3B model runs in 36 ms on a single B200 GPU, making it 7.6× faster than CF++. Our 14B model runs 10.6× faster than the fastest open-source AR model of the same parameter count and achieves a further 3.5× speedup when distributed across eight B200 GPUs.

\subsection{Removing the VAE Decoder}
\label{sec:decoder}

We first target the VAE decoder since it accounts for more than half of the end-to-end inference latency and alone exceeds our real-time budget 3×.
The VAE decoder only has 73M parameters (compared to the DiT's 1.3B) but repeatedly upsamples its features in space and time and applies dense 3D convolutions at each resolution. For example, a tensor containing four 480×832 frames requires convolution at approximately 1.6 million spatiotemporal positions.

\input{figures/vae_ablation}

In light of recent advances in pixel diffusion models, we question whether such a decoder is necessary when the DiT already produces rich and spatially contextualized features. For instance, JiT~\citep{li2025jit} produces sharp images by mapping between image patches and transformer features using only linear projections. 
Thus, we first attempt replacing the VAE with a single linear layer and train it with an MSE loss against the VAE decoded output, keeping the DiT frozen. We found that the outputs were blurry yet recognizable (Fig.~\ref{fig:vae_ablation}). 

We then increase the size of the pixel decoder, until arriving at our final architecture consisting of three causal 3D ConvNeXt-style blocks~\citep{liu2022convnet}. Each block uses a causal 3×3×3 depthwise convolution followed by a channel-mixing MLP. All blocks operate only on the transformer's low-resolution 21×30×52 token grid. A final linear layer converts each token into a 4×16×16 RGB patch, yielding an 84-frame, 480×832 video. We discard the first three frames following the conventions of the Wan VAE, which does not temporally upsample the first frame. The total number of parameters is comparable (63.9M for ours vs.\ 73.3M), yet two orders of magnitude faster (1 ms vs.\ 147 ms). 

We train the pixel decoder using spatial and temporal reconstruction objectives. Let $\hat{y}_{\mathrm{pix}}$ denote the RGB video predicted by the ConvNeXt decoder, and let $y_{\mathrm{VAE}}$ denote the output obtained by decoding the same final latent $x_0$ with the frozen VAE.
We optimize
\begin{equation}
\mathcal{L}_{\mathrm{rec}}(\hat{y}_{\mathrm{pix}},y_{\mathrm{VAE}})
=\lambda_{1}\mathcal{L}_{1}
+\lambda_{\mathrm{MSE}}\mathcal{L}_{\mathrm{MSE}}
+\lambda_{\mathrm{grad}}\mathcal{L}_{\mathrm{grad}}
+\lambda_{\mathrm{temp}}\mathcal{L}_{\mathrm{temp}}
+\lambda_{\mathrm{motion}}\mathcal{L}_{\mathrm{motion}}.
\label{eq:pixel_reconstruction}
\end{equation}
Here, $\lambda_{1}=0.5$, $\lambda_{\mathrm{MSE}}=5.0$,
$\lambda_{\mathrm{grad}}=\lambda_{\mathrm{temp}}=0.05$, and
$\lambda_{\mathrm{motion}}=0.5$.
All five terms are computed between $\hat{y}_{\mathrm{pix}}$ and $y_{\mathrm{VAE}}$.
The $\mathcal{L}_{1}$ and $\mathcal{L}_{\mathrm{MSE}}$ terms enforce pixel-level
fidelity, while $\mathcal{L}_{\mathrm{grad}}$ matches spatial image gradients
and $\mathcal{L}_{\mathrm{temp}}$ matches frame-to-frame changes.
The motion-weighted term $\mathcal{L}_{\mathrm{motion}}$ emphasizes
reconstruction errors in regions exhibiting motion in the VAE target.
Complete definitions of the loss terms are provided in the supplement.

In addition to reducing latency, our pixel decoder is less memory-intensive to optimize. For a batch-size-one, 81-frame
video at 832×480 resolution, a VAE decoder update with BF16 precision and activation checkpointing requires approximately 68\,GB of peak GPU memory, compared with approximately 5\,GB for our pixel decoder. This makes it practical to optimize the decoder directly in pixel space with adversarial losses to improve sharpness and realism. 

We curated an internal dataset of 10K real videos and used a frozen DINOv3~\citep{simeoni2025dinov3} ViT-S/16 backbone as our discriminator. 
On each of layers 4, 8, and 12, we train two lightweight token-level discriminator heads: a spatial classifier and a temporal classifier that receives both the current patch feature and its change from the preceding frame.
Let $\mathcal{H}$ denote the six discriminator heads, let $y$ denote a real video crop, and let $\hat{y}$ denote a generated crop. We use the following relativistic
softplus discriminator objective~\citep{jolicoeur2018relativistic}:
\begin{equation}
\mathcal{L}_{D}
=\frac{1}{|\mathcal{H}|}\sum_{h\in\mathcal{H}}
\left\langle
\operatorname{softplus}\!\left(D_h(\hat{y})-D_h(y)\right)
+\lambda_{\mathrm{center}}\left[D_h(y)+D_h(\hat{y})\right]^2
\right\rangle
+\lambda_{\mathrm{slope}}(R_1+R_2),
\end{equation}
where $\langle\cdot\rangle$ denotes averaging over clips, frames, and patch tokens, $\lambda_{\mathrm{center}}=0.01$, and $\lambda_{\mathrm{slope}}=0.25$. The terms $R_1$
and $R_2$ are finite-difference slope penalties on real and generated inputs, respectively~\citep{afm}. We estimate them on 25\% of the clips by adding Gaussian RGB perturbations with standard deviation $\epsilon=0.01$.

The generator consists of the frozen DiT and the trainable ConvNeXt pixel decoder. We continue using the reconstruction objective (Eq.~\ref{eq:pixel_reconstruction}) with the reverse relativistic comparison:
\begin{equation}
\mathcal{L}_{\mathrm{adv}}=\frac{1}{|\mathcal{H}|}\sum_{h\in\mathcal{H}}\left\langle\operatorname{softplus}\left(-\left[D_h(\hat{y})-\operatorname{sg}\!
\left(D_h(y)\right)\right]\right)\right\rangle,
\end{equation} where $\operatorname{sg}$ denotes stop-gradient. We cap the adversarial RGB-gradient magnitude relative to the reconstruction gradient for stable optimization.
We also tried finetuning the DiT with low learning rate, but observed training instability.

As seen in Fig.~\ref{fig:vae_ablation}, our pixel decoder faithfully reconstructs thin structures like hair and raindrops. However,
unlike the Wan VAE decoder, which can be shared across models of the same family (\eg 1.3B and 14B variants), our pixel decoder must be finetuned for each individual model. Our main takeaway is that we can turn pixel decoding into a lightweight readout rather than a general-purpose reconstruction problem.

\subsection{Faster KV Cache Updates}

\label{sec:kvcache}

Autoregressive video diffusion models attend to earlier frames through a KV cache~\citep{yin2024causvid, huang2026selfforcing}. 
Each DiT layer maintains its own cache, which holds an attention sink plus a sliding window of recent entries. During each generation, the final denoising pass leaves the current chunk’s cache entries computed from a noisy input. The baselines therefore run an additional DiT pass on the predicted clean latent $\hat{\vz}_i$ at $t=0$ and replace those entries with its keys and values (the KV cache write in Fig.~\ref{fig:pipeline}). They discard the pass’s output prediction.

In a two-step diffusion setup, KV refinement accounts for one-third of DiT evaluations. We first investigate whether refinement is necessary. Naively removing the refinement leads to substantial degradation and training the model to use noisy cached features still produces noticeable artifacts (see supplement for visualizations).
However, we hypothesize that a lightweight network can replace the expensive DiT pass. 
We design a KV refinement network with a shared feature extractor and per-layer KV-repair heads. The feature extractor takes $\hat{\vz}_i$ and applies the DiT's frozen patch embedding followed by four lightweight transformer blocks trained from scratch. Each repair head takes these output features and its layer's noisy keys and values to predict residual corrections.

During training, we freeze the DiT and use the keys and values produced by the original DiT refinement pass to supervise residual corrections to the noisy cache. We train our network with two losses: a direct KV reconstruction loss combining cosine distance and relative $\ell_2$ error, and an attention-output loss that compares the outputs obtained when subsequent chunks attend to the predicted versus target cache.
The network performs comparably to the original refinement pass in terms of quantitative and qualitative comparisons, while taking approximately 2 ms for the 1.3B model and 6 ms for the 14B model on a B200 GPU, around 20× and 12× faster respectively than the baselines' refinement. We provide details in the supplement.

\input{tables/t2v_performance}

\input{figures/qual_comparison}
\subsection{Systems-level Optimizations}
\label{sec:sysopt}

After reducing non-denoising overhead, DiT evaluations remain the main bottleneck to real-time generation. ``Free'' optimizations, including faster attention backends and \texttt{torch.compile()}, are already accounted for in the baseline profiling (Fig.~\ref{fig:pipeline}). 
First, we observe that the repeated structure of DiT evaluations offers opportunities to reuse computation. 
We maintain persistent KV-cache buffers and write new entries directly into their designated slots; precompute and store RoPE tables, text and timestep conditions; and fuse QKV projections into one larger matrix multiplication. 

Second, we find that per-frame AR generation can provide too few query tokens to fully utilize the GPU. We therefore split the KV sequence into segments processed in parallel, then combine their attention outputs using a softmax merge. Concretely, for a full KV cache of length 32,760, we found that splitting it into five segments of 6,552 each led to a 13\% latency reduction.
Third, we reduce the overhead of launching and coordinating GPU work via CUDA graphs. 
Together, these optimizations bring the DiT evaluation latency from 41.6 ms to 16.4 ms for the 1.3B model on one B200 GPU at the full 32,760-token KV prefix and from 89.2 to 75.6 ms for the 14B model, corresponding to approximately 2.5× and 1.2× speedups. 

Next, we distribute computation across 8× B200 GPUs. We experimented with tensor parallelism, attention-head parallelism, KV sharding, joint query–KV partitioning, and context parallelism (CP). 
Among them, only CP reduced computation sufficiently to justify communication overhead. 
CP partitions query tokens across GPUs and distributes both attention and feed-forward computation, requiring only one exchange of computed keys and values per transformer layer. We implement these exchanges using CUDA-graph-compatible NVLink symmetric-memory gathers. This configuration reduces DiT latency from 16.4 to 5.8 ms for the 1.3B model and from 75.6 to 20.3 ms for the 14B model, corresponding to 2.8× and 3.7× speedups over their optimized single-GPU implementations.

%% file: figures/vae_ablation.tex
\begin{figure}
    \centering
    \includegraphics[width=\linewidth]{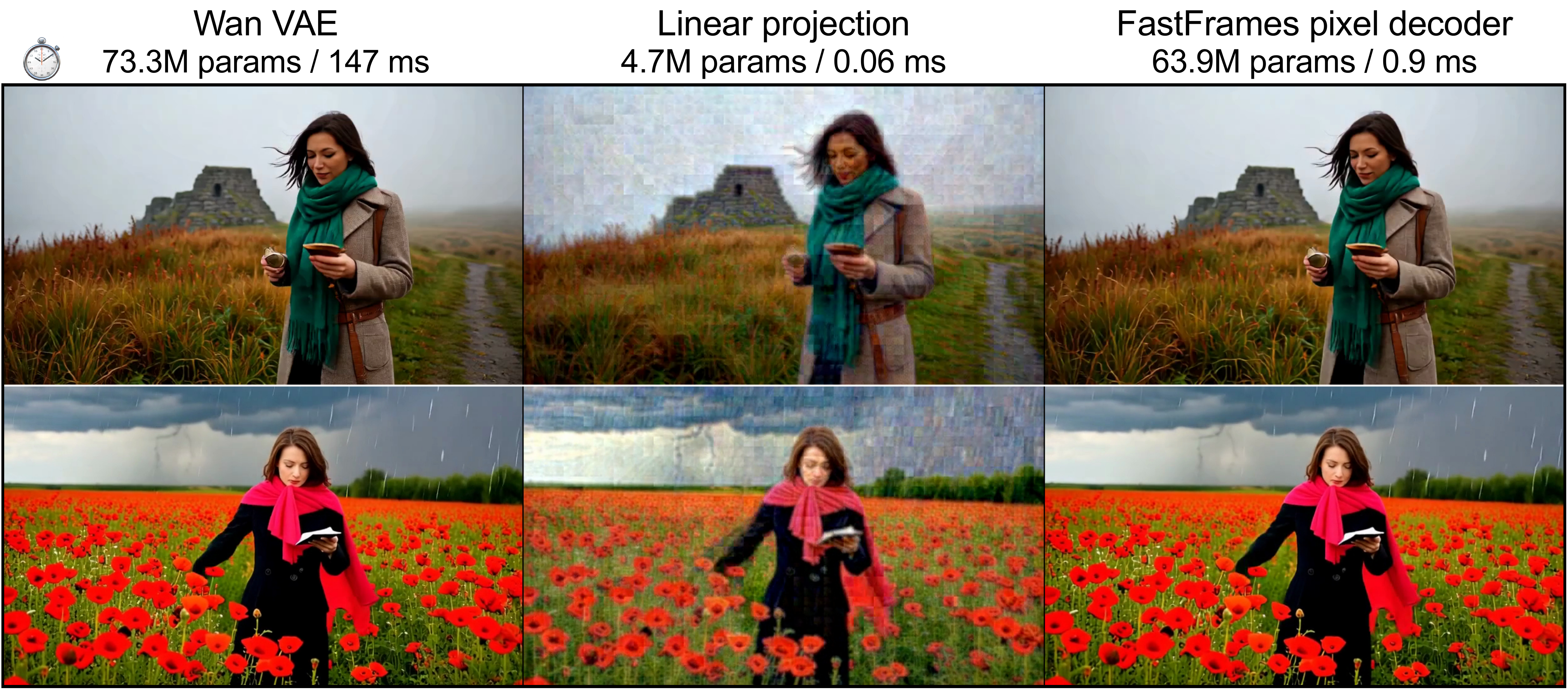}
 
    \caption{
    \textbf{Comparing pixel decoders.}
    Our pixel decoder (right) renders thin structures and details like hair and raindrops while being two orders of magnitude faster than Wan's VAE (left). A single linear projection (middle) cannot render sharp outputs, but is sufficient for recognizing the scene and suggests that pixel rendering does not require an expensive decoder.
    }
    \label{fig:vae_ablation}
    
\end{figure}

%% file: tables/t2v_performance.tex
\begin{table*}[t]
\centering
\small
\setlength{\tabcolsep}{4.5pt}
\caption{\textbf{Quantitative comparisons in terms of end-to-end latency and VBench scores.} All methods use the same evaluation prompts and seeds, run on one B200 GPU unless otherwise indicated, and are evaluated at approximately 480p using their officially supported dimensions. \colorbox{latencyFast}{Green} indicates latency that meets the interactive target ($\leq 50$ ms); \colorbox{latencySubsecond}{yellow} indicates sub-second latency ($50 < \text{latency} < 1{,}000$ ms).
}
\label{tab:quantitative}
\begin{tabular}{@{}l r r c c c@{}}
    \toprule
    & & & \multicolumn{3}{c}{VBench $\uparrow$} \\
    \cmidrule(lr){4-6}
    Model & Size (B) & Latency (ms) $\downarrow$ &
    Total & Quality & Semantic \\
    \midrule
    \multicolumn{6}{l}{\textit{Bidirectional Video Models}} \\
    Wan2.1~\citep{wan2025wan}                & 1.3  & 30,444    & 83.80 & 84.56 & 80.73 \\
    Wan2.2~\citep{wan2025wan}                & 14.3 & 117,734   & 84.65 & 85.41 & 81.60 \\
    MiniMax-H3~\citep{minimax2026h3}            & 33.1 & 38,454    & 84.43 & 85.17 & 81.47 \\
    HunyuanVideo1.5~\citep{wu2025hunyuanvideo15} & 8.3  & 1,555,508 & 83.18 & 84.14 & 79.35 \\
    LTX2.5~\citep{hacohen2026ltx2} & 19.0 & 11,264    & 83.62 & 84.78 & 78.96 \\
    \midrule
    \multicolumn{6}{l}{\textit{Autoregressive Video Models}} \\
    MAGI1.1 Distilled~\citep{ai2025magi1autoregressivevideogeneration}  & 23.9 & 19,736       & 82.04 & 83.40 & 76.60 \\
    Helios Distilled~\citep{helios}      & 14.3 & 2,544        & 83.43 & 84.53 & 79.03 \\
    Krea-Realtime~\citep{krea_realtime_14b}         & 14.3 & 1,675        & 84.42 & 85.08 & 81.81 \\
    LongLive2.0~\citep{chen2026longlive}   & 5.0  & \latsub{505} & 83.65 & 84.54 & 80.09 \\
    CausVid~\citep{yin2024causvid}               & 1.3  & \latsub{662} & 83.18 & 84.15 & 79.31 \\
    Self-Forcing\textsuperscript{*}~\citep{huang2026selfforcing}\          & 1.3  & \latsub{717} & 84.10 & 84.84 & 81.13 \\
    Causal Forcing++~\citep{zhao2026causal}      & 1.3  & \latsub{272} & 84.11 & 85.09 & 80.20 \\
    \addlinespace
    \textbf{\name-S}         & 1.3  & \latfast{36} & 84.23 & 85.30 & 79.92 \\
    \textbf{\name-S (8 GPU)} & 1.3  & \latfast{14} & 84.23 & 85.30 & 79.92 \\
    \textbf{\name-L}         & 14.3 & \latsub{158} & 84.44 & 85.48 & 80.29 \\
    \textbf{\name-L (8 GPU)} & 14.3 & \latfast{45} & 84.44 & 85.48 & 80.29 \\
    \bottomrule
\end{tabular}
\par\vspace{3pt}
\begin{minipage}{\linewidth}
    \scriptsize
    \raggedright
    \textsuperscript{*}Self-Forcing reports latency averaged over a sequence, including cheaper early frames as the KV cache fills. We measure steady-state latency with a full cache. See the supplement for details.
\end{minipage}
\end{table*}

%% file: figures/qual_comparison.tex
\begin{figure}[t]
\centering
\includegraphics[width=0.94\textwidth]{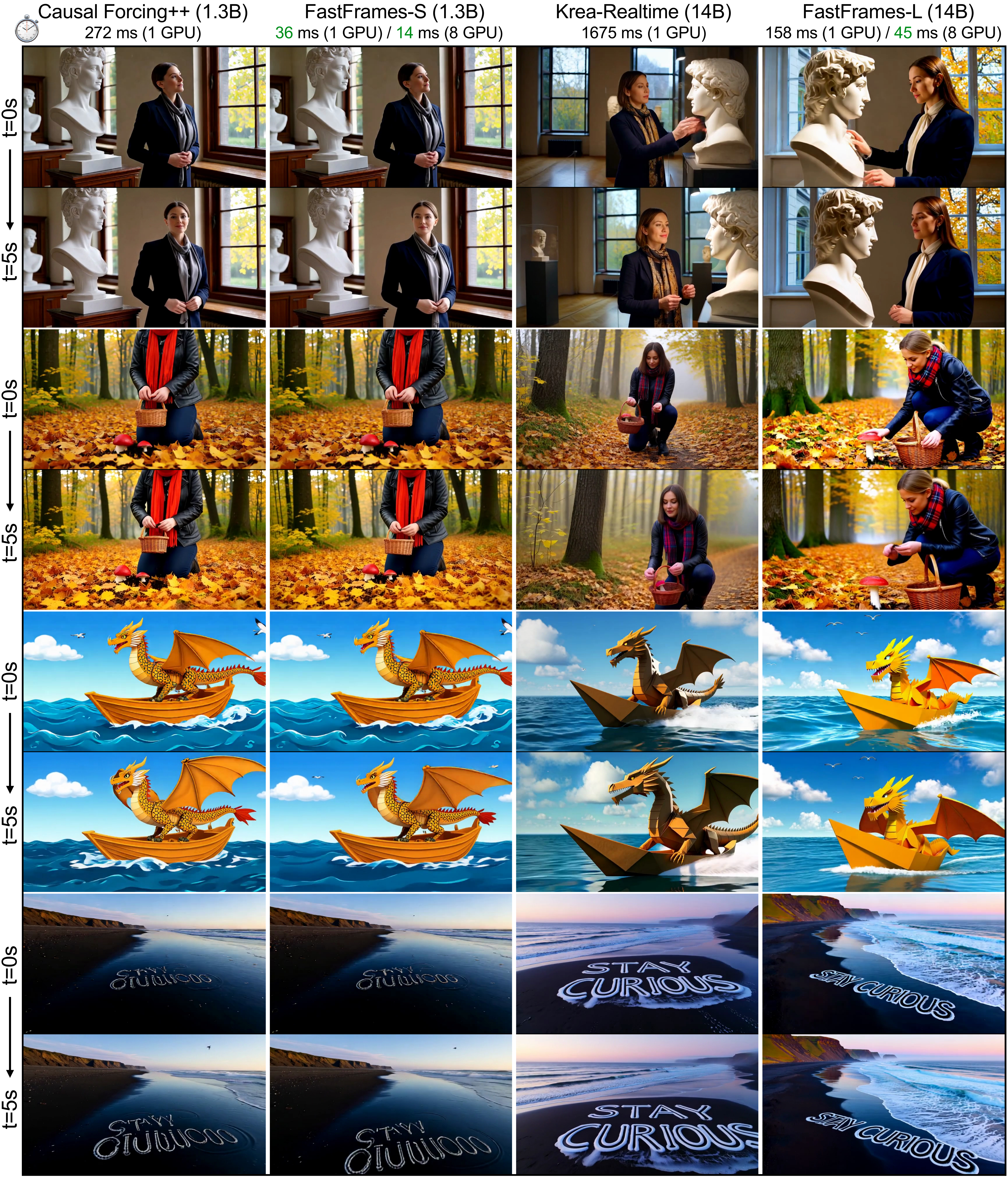}

\caption{\textbf{Qualitative comparisons.}
We compare \name-S with Causal Forcing++ at 1.3B parameters and \name-L with Krea-Realtime at 14B. Each example shows frames at $0$ and $5$ seconds.
\name-S achieves comparable visual sharpness to Causal Forcing++, such as the details in the sculpture (first example) and leaves (second example), while being nearly 8× faster.
\name-L achieves comparable visual realism to Krea-Realtime, including composition of ``paper origami dragon'' (third), and legible text of ``stay curious'' (fourth) while being 10× faster.
See our \href{https://apple-aiml-research.github.io/ml-fastframes/}{website} for video playback. 
}

\label{fig:comparison}
\end{figure}

%% file: sections/04_results.tex
\section{Experiments}
\label{sec:exp}
\textbf{Models and baselines.}
We build \name on the Wan 2.1 backbone at two model sizes: \name-S with 1.3B parameters and \name-L with 14B.
We initialize \name-S from the two-step Causal Forcing++ (CF++) checkpoint and finetune it using our method. We freeze the DiT and train our pixel decoder and KV refinement network against their respective targets. \name-L required starting from the base bidirectional model. Following the training recipe in CF++, we converted Wan 2.1 14B into an autoregressive model, distilled it into a few-step model, and trained it with DMD. Both \name-S and \name-L perform two-step diffusion sampling and generate one latent frame per autoregressive step.

\textbf{Quantitative comparisons.}
Tab.~\ref{tab:quantitative} provides a quantitative comparison against bidirectional video generators and autoregressive models. The bidirectional models provide a broader reference for generation quality and latency, while the autoregressive models provide direct comparisons. We report model size, latency, and VBench~\citep{huang2023vbench} scores. All models are evaluated using the same VBench prompts and seeds.
We report end-to-end latency and compute RPS as its reciprocal in seconds, following the definitions in the introduction. For AR models, one response corresponds to an autoregressive generation unit; for bidirectional models, it corresponds to a complete generated clip. Consequently, RPS does not credit a method for producing multiple frames within one generation unit. We account for diffusion sampling, KV cache updates where applicable, and pixel rendering. For AR models, we measure latency at steady state after warm-up and with a full KV cache. For bidirectional models, we let the model warm up and then generate the configured native clip lengths. All models are evaluated on a B200 GPU unless otherwise specified, at approximately 480p using the model's officially supported resolutions.

We find that \name matches corresponding baselines with the same Wan 2.1 backbone (Krea-Realtime vs.\ \name-L, CF++ vs.\ \name-S). Between \name-S and CF++, which use the same DiT weights, we attribute the higher visual quality to end-to-end adversarial training in pixel space, which boosts realism and sharpness.

None of the baselines meet the 50 ms latency target required for interactive applications. The fastest baseline, CF++, reaches 3.68 RPS at 272 ms, falling 5.4× short of the 20 RPS target. In contrast, \name-S attains a latency of 36 ms (27.8 RPS) on a single B200 GPU and \name-L attains 45 ms (22.2 RPS) on an 8-GPU node. \name-S on 8 GPUs reaches 14 ms (71.4 RPS), while \name-L on a single GPU reaches 158 ms (6.33 RPS).

\textbf{Qualitative comparisons.}
Fig.~\ref{fig:comparison} shows a side-by-side comparison between \name-S, \name-L, and their respective baselines. \name-S maintains visual sharpness relative to CF++, capturing fine details like sculptures and leaves while being nearly 8× faster. 
\mbox{\name-L} achieves realism, prompt adherence, composition, and text rendering capabilities comparable to Krea-Realtime while being over 10× faster.

Fig.~\ref{fig:comparison} also illustrates the improved capabilities of the 14B models relative to their 1.3B counterparts. The larger models produce more realistic lighting and shadows in the first two examples, better capture the folded-paper appearance of the origami dragon in the third, and render more legible text in the fourth.

\input{tables/ablation_components}
\textbf{Ablation study.}
Tab.~\ref{tab:ablation_components} ablates the two lightweight networks of \name-S. Starting from the Causal Forcing++ two-step DiT, we replace the VAE with our pixel decoder (Sec.~\ref{sec:decoder}), the KV-refinement DiT evaluation with our KV refinement network (Sec.~\ref{sec:kvcache}), or both. VBench uses the same prompts and seeds as Tab.~\ref{tab:quantitative}. To isolate the two components, every row uses our optimized DiT runtime (Sec.~\ref{sec:sysopt}), and latency is the sum of measured per-component timings. The first row is therefore faster than the 272 ms reported for Causal Forcing++ in Tab.~\ref{tab:quantitative}, which uses its original runtime.

Both replacements preserve quality: all four configurations are within 0.12 VBench total points, comparable to the standard deviation across seeds ($\pm$ in Tab.~\ref{tab:ablation_components}). In the three-seed rerun behind these error bars, no configuration's total differs significantly from the full model (paired bootstrap over prompts).

Beyond VBench, our KV refinement network yields steadier video than the DiT evaluation it replaces: with the DiT evaluation instead, optical-flow warping error increases by about 10\% and FVD by about 29 (Tab.~\ref{tab:ablation_components}), consistently across seed sets. We attribute this to specialization: our network is trained solely for cache repair, whereas the DiT evaluation reuses the general-purpose generator.

VBench does not capture everything. Learned perceptual metrics prefer the VAE, mainly human-preference reward (VideoReward) and per-frame quality consistency (VMBench motion smoothness), as expected for a decoder distilled from the VAE that runs 147$\times$ faster. Gaps on general video quality assessors are small (Q-Align: 4.84 vs.\ 4.90 out of 5; DOVER++: within 1\%), and our decoder even has slightly lower FVD than the VAE (Tab.~\ref{tab:ablation_components}). Because generation is unchanged, the VAE remains a drop-in decoder when latency is less critical.

\textbf{Interactive applications.}
\label{sec:app}
We showcase \name in interactive settings by accelerating camera-conditioned and physics-conditioned video generation: LingBot~\citep{lingbot-world-v2}, a family of camera-conditioned video models with 1.3B and 14B variants, and RealWonder~\citep{liu2026realwonder}, a 1.3B physics-conditioned video model. 
Our techniques transfer directly to these off-the-shelf models and yield large speedups, even over official code that already includes optimizations such as Ulysses sequence parallelism~\citep{jacobs2023deepspeedulysses}. Fig.~\ref{fig:interaction} summarizes the latency profiles.

On LingBot 1.3B, we reduce latency from 842 ms (the fastest official configuration, 2 GPUs) to 52 ms on 8 GPUs, a 16× speedup. On RealWonder, latency drops from 1141 ms (official, 1 GPU) to 51 ms (22×), so both models approximately reach our target interactive latency. On LingBot 14B, latency drops from 1373 ms to 233 ms on 8 GPUs (5.9×). 
These models remain slower than our T2V models because they require more diffusion sampling steps, predict multiple latent frames per autoregressive step, and need extra DiT computation to process additional conditioning.

We show a side-by-side comparison between the interactive demos on the \href{https://apple-aiml-research.github.io/ml-fastframes/}{project website}; the original models exhibit noticeable lag while our optimized versions of LingBot 1.3B and RealWonder respond to user inputs fluidly. 
\input{figures/lingbot_profile}

\input{figures/fig_user_study}
\input{tables/user_study}
\textbf{User study.}
We evaluate interactive control with a user study on LingBot 1.3B (Fig.~\ref{fig:interaction}), comparing its official implementation with our accelerated version. Rather than a passive side-by-side video comparison, which can only compare visual quality, we design trials that put participants in the loop and allow them to experience the responsiveness and controllability of each model, interactively.

Specifically, the participant is given a starting image, a target image, and six controls: move forward, backward, left and right (W/S/A/D) and pan left and right ($\leftarrow$/$\rightarrow$). The goal is to control the video model to move from the first image to the target image within at most 30 seconds, and match the screen output to the target image as closely as possible (Fig.~\ref{fig:user_study}, left).

A concrete goal, like a video game, allows participants to control both models in the same way and gives an objective measure of how close they get (LPIPS to the target). The study thus tests whether our lower latency translates into a perceptibly more responsive and controllable experience without a loss in photorealism.

In each trial, the participant controls both our method and the baseline, one after the other. The models are labeled only Video A and Video B, and each model goes first in half of each participant's trials. Afterwards, the participant watches replays of both attempts side by side with the target (Fig.~\ref{fig:user_study}, right) and answers four questions: which video model is (1) more responsive, (2) more photorealistic, (3) more controllable, and (4) more suitable for interactive applications like gaming? Each question has three choices: Video A, Video B or Equal.

We received 74 responses. Participants judged our model more responsive in 72\% of trials, more controllable in 47\%, and more suitable for interactive applications in 49\%, while most rated the two as equally photorealistic (54\%; Tab.~\ref{tab:user_study}). Both models get similarly close to the target: the mean minimum LPIPS is 0.46 for ours and 0.44 for the official implementation.

%% file: tables/ablation_components.tex
\begin{table}[t]
\centering
\small
\setlength{\tabcolsep}{5pt}
\caption{\textbf{Ablation of \name-S's lightweight networks} on the 1.3B Causal Forcing++ two-step DiT. Each row renders pixels with the VAE or our pixel decoder, and updates the KV cache with an extra DiT evaluation or our KV refinement network. Latency is per frame on one B200 GPU at 480p, with our optimized DiT runtime in every row. VBench uses the evaluation of Tab.~\ref{tab:quantitative}. $\Delta$FVD (I3D, 16 consecutive frames, against 6,220 real videos) is relative to the full model ($726.5\pm2.4$). All $\pm$ are standard deviations over three seed sets in a separate controlled rerun.}
\label{tab:ablation_components}
\begin{tabular}{@{}l l r c c@{}}
    \toprule
    Pixel & KV & Latency & VBench & $\Delta$FVD \\
    rendering & update & (ms) $\downarrow$ & Total $\uparrow$ & $\downarrow$ \\
    \midrule
    VAE  & DiT  & 196 & 84.11{\scriptsize\,$\pm$\,0.14} & $+$34.3{\scriptsize\,$\pm$\,8.2} \\
    VAE  & Ours & 182 & 84.15{\scriptsize\,$\pm$\,0.05} & $+$6.5{\scriptsize\,$\pm$\,0.5} \\
    Ours & DiT  & 50  & 84.16{\scriptsize\,$\pm$\,0.16} & $+$29.0{\scriptsize\,$\pm$\,7.3} \\
    Ours & Ours & 36  & 84.23{\scriptsize\,$\pm$\,0.05} & 0 \\
    \bottomrule
\end{tabular}
\end{table}

%% file: figures/lingbot_profile.tex
\begin{figure*}[t]
  \centering

\begin{minipage}[c]{0.65\textwidth}
  \centering
  \includegraphics[width=\linewidth]{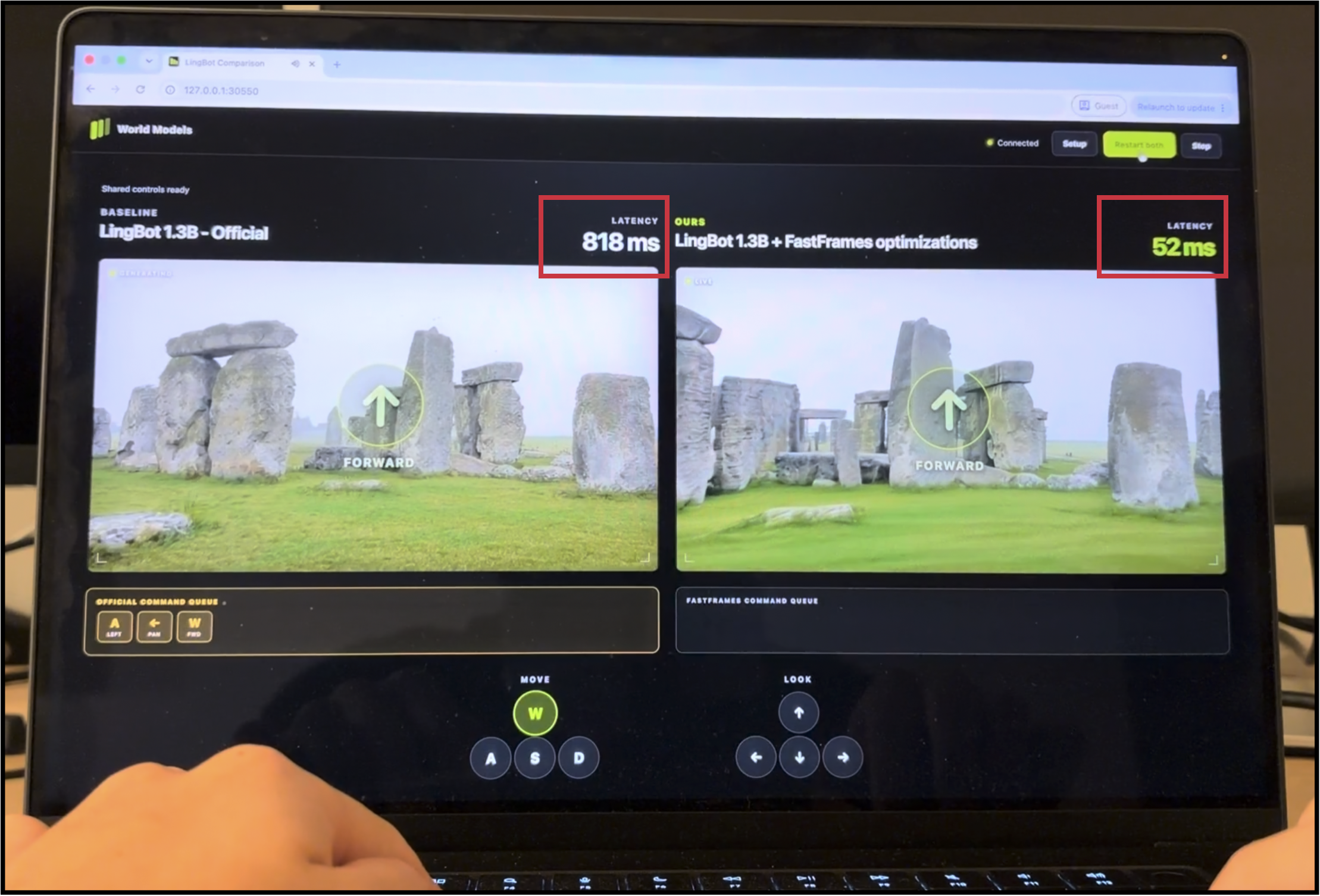}
\end{minipage}%
\hspace{0.02\textwidth}%
\begin{minipage}[c]{0.32\textwidth}
  \raggedright
  \fontsize{8.3pt}{9.8pt}\selectfont 
  \setlength{\tabcolsep}{3.5pt}
  \renewcommand{\arraystretch}{0.95}
  \begin{tabular*}{\linewidth}{@{\extracolsep{\fill}}lr@{}}
      \toprule
      Configuration & \makebox[0pt][r]{Latency (ms)} \\
      \midrule
      \multicolumn{2}{@{}l}{\textbf{LingBot 1.3B}} \\
      Official inference - 1 GPU & 1085 \\
      Official + Ulysses - 2 GPU &  842 \\
      Official + Ulysses - 4 GPU &  865 \\
      W/ \name\ - 1 GPU &  224 \\
      W/ \name\ - 8 GPU &   52 \\
      \midrule
      \multicolumn{2}{@{}l}{\textbf{LingBot 14B}} \\
      Official inference - 1 GPU & 3862 \\
      Official + Ulysses - 8 GPU & 1373 \\
      W/ \name\ - 1 GPU & 1263 \\
      W/ \name\ - 8 GPU &  233 \\
      \midrule
      \multicolumn{2}{@{}l}{\textbf{RealWonder 1.3B}} \\
      Official inference - 1 GPU & 1141 \\
      W/ \name\ - 1 GPU &  153 \\
      W/ \name\ - 8 GPU &   51 \\
      \bottomrule
    \end{tabular*}
  \end{minipage}

  \caption{
    \textbf{Interactive generation and latency results.}
    Left: example of interaction with LingBot's official implementation vs.\ with \name optimizations. Given a single keypress, \name is 16× faster to respond.
    Right: generation latency of officially released implementations and our optimized implementations.
    We measure steady-state latency on NVIDIA B200 GPUs at 832×480 resolution with a full KV cache and include distributed optimizations such as Ulysses where officially supported.
    Times include diffusion sampling (four steps for LingBot and three for RealWonder), action conditioning, KV-cache updates, and pixel decoding.
    Our optimizations provide speedups of 3× to 22×.
  }
  \label{fig:interaction}
  
\end{figure*}

%% file: figures/fig_user_study.tex
\begin{figure}[t]
\centering
\includegraphics[width=0.57\linewidth]{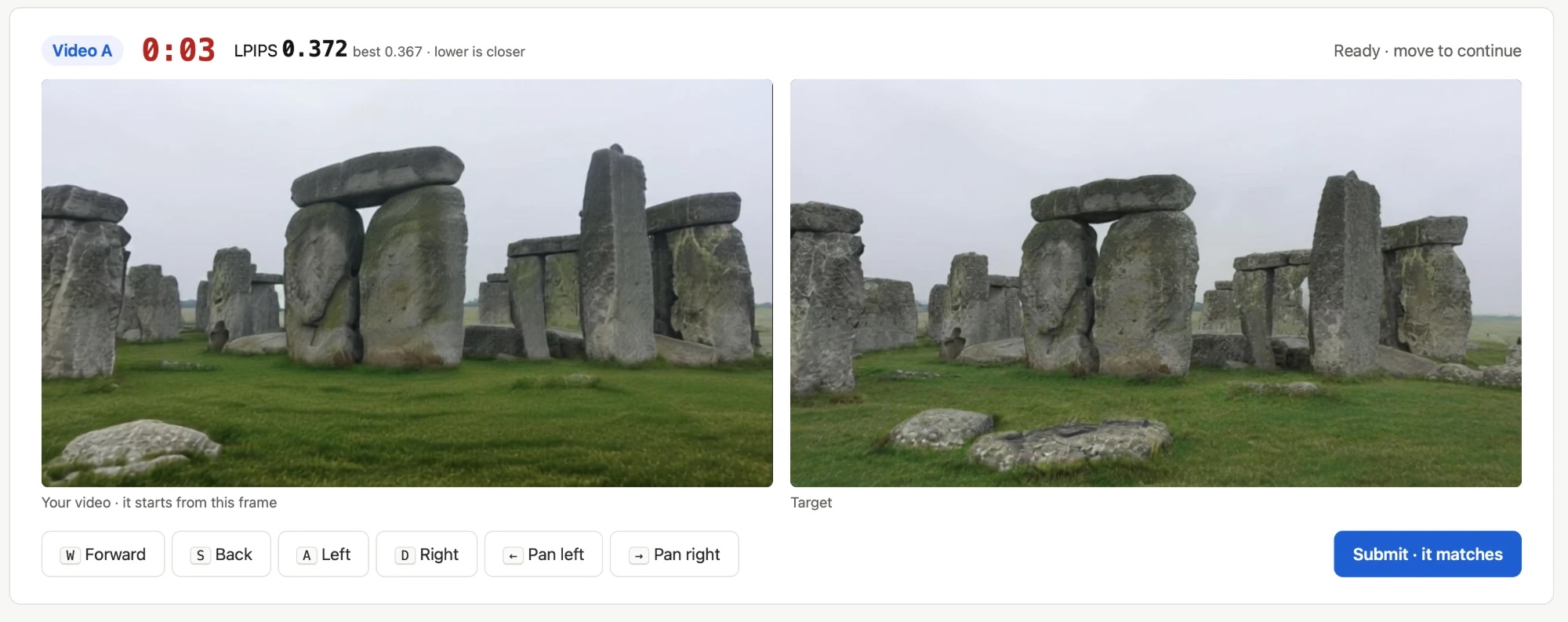}\hfill
\includegraphics[width=0.415\linewidth]{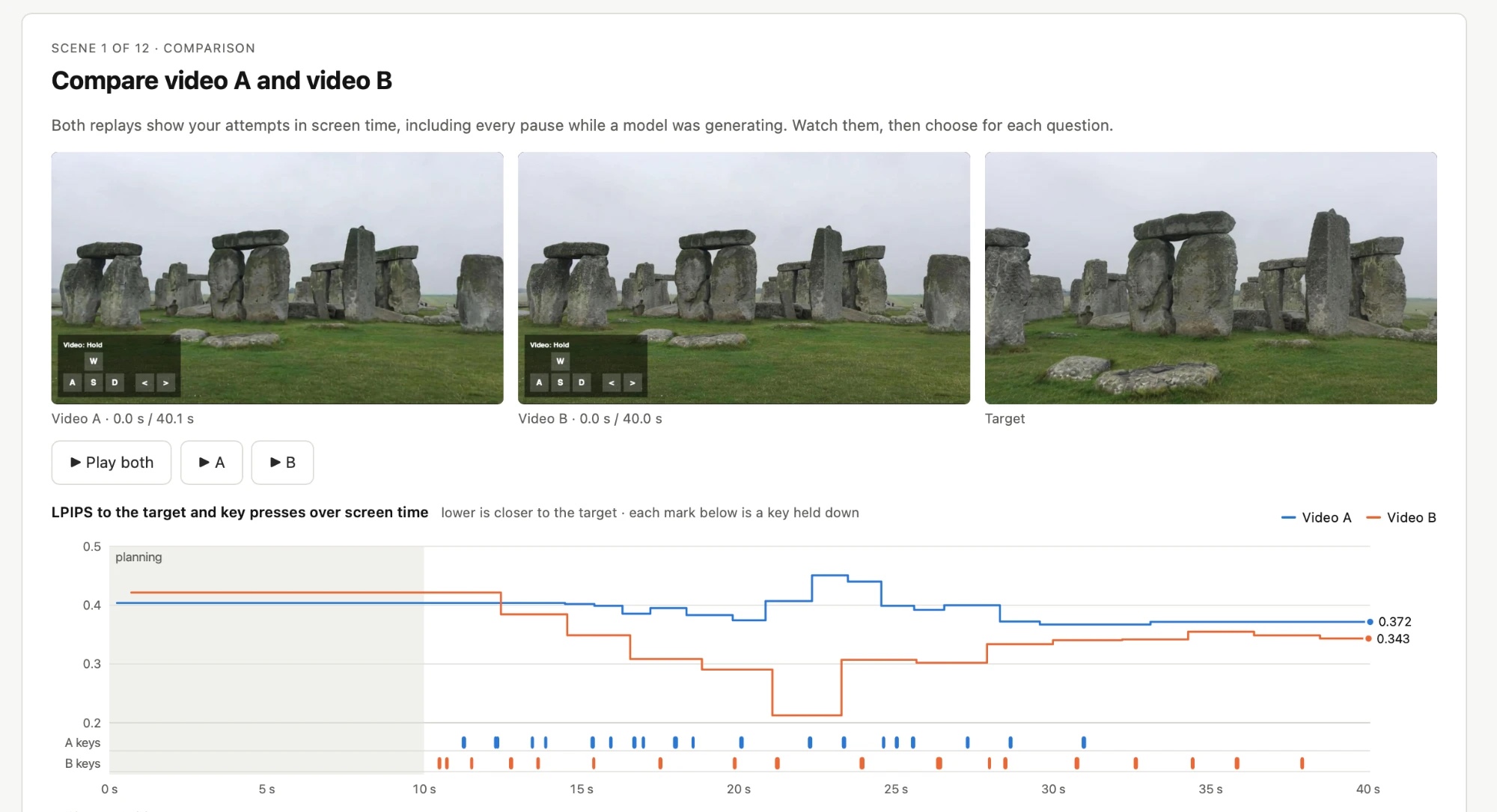}
\caption{\textbf{User study interface.} \emph{Left:} steering screen. As in a video game, the goal is to move the camera until the video matches the target within a time limit. The left image is the live output of the model being controlled (here Video A), and the right image is the target. The red timer shows the remaining steering time: every attempt gets the same 30 s, so a slower model completes fewer actions. LPIPS is the perceptual distance between the current frame and the target (lower is closer), shown with the best value reached so far, so the participant gets objective feedback on whether each action brings them closer. The buttons below mirror the keyboard controls. \emph{Right:} comparison screen, shown after both attempts. It replays the participant's attempts with Video A and Video B next to the target, plots each attempt's LPIPS to the target and its key presses over time, and is followed by the four questions.}
\label{fig:user_study}
\end{figure}

%% file: tables/user_study.tex
\begin{table}[t]
\centering
\small
\setlength{\tabcolsep}{5pt}
\caption{\textbf{User study on LingBot 1.3B.} Percentage of trials in which participants preferred \name, the official implementation, or neither.}
\label{tab:user_study}
\begin{tabular}{@{}l c c c@{}}
    \toprule
    Which model is more \ldots & \name & Official & Equal \\
    \midrule
    responsive               & \textbf{72\%} & 9\%  & 19\% \\
    photorealistic           & 19\% & 27\% & \textbf{54\%} \\
    controllable             & \textbf{47\%} & 24\% & 28\% \\
    suitable for gaming      & \textbf{49\%} & 22\% & 30\% \\
    \bottomrule
\end{tabular}
\end{table}

%% file: sections/05_conclusion.tex
\section{Conclusion}
\label{sec:conclusion}

We presented \name, a family of autoregressive video diffusion models that achieves interactive latency by rethinking components of the inference pipeline often treated as fixed implementation details. Both our 1.3B and 14B variants achieve generation quality comparable to or better than their respective baselines.

While our evaluation focuses on text-to-video generation at 480p on high-end GPUs, important directions for future work include adding action conditioning without increasing latency, scaling to higher resolutions, and supporting lighter hardware form factors. Improving long-term consistency and physical plausibility will further broaden the utility of video world models.

%% file: sections/x_supp.tex
\section*{Appendix}
We organize the appendix into the following sections: (A) Limitations and future work, (B) Additional latency profiling, (C) Architecture and training details of pixel decoder, (D) Architecture and training details of KV refinement network, and (E) Adaptation details of interactive applications.

We encourage readers to view the videos on our \href{https://apple-aiml-research.github.io/ml-fastframes/}{website}.

\appendix
\section{Limitations and Future Work}
\textbf{Resolution.} Most baselines and base video models use 480p resolution, which we finetune from. Higher resolution like 720p or even 4K is desired for a variety of applications like gaming and content creation. In the next section, we find that increasing resolution to 720p increases latency significantly; further optimization is required to bring high-res video generation into the interactive regime.

\textbf{Accessible hardware.} We use B200 GPUs for our latency measurements in the main paper. While we assume compute will continue to become cheaper and more accessible, being able to run interactive video models on-device is an exciting possibility.

\textbf{Action conditions designed for lower latency.} We applied \name to off-the-shelf autoregressive models that were slower than their T2V counterparts. Training base models with lower latency in mind, such as concatenating camera poses in the channel dimension rather than the sequence dimension, without compromising quality, is essential for reducing latency.

\section{Additional Latency Profiling}
\input{tables/additional_profiling}
In Tab.~\ref{tab:lingbot_latency}, we include detailed profiling of each operation on B200 and H100 GPUs, and at 480p and 720p resolution. In short, \name-S runs at 36 ms at 480p and 127 ms at 720p on a B200 GPU, and \name-L runs at 45 ms at 480p and 116 ms at 720p on 8 B200 GPUs. Savings from our lightweight networks are more noticeable at 720p. H100s are approximately 25\% to 50\% slower depending on the configuration. 

Some baselines like Self-Forcing~\citep{huang2026selfforcing} report latency in their paper. However, we did not find detailed descriptions or breakdowns of those numbers and can only infer their profiling from source code. We found that Self-Forcing measured latency by generating a video sequence of N latent chunks (where each chunk $\geq$ 1 latent frame(s)), then dividing the total time by the number of chunks. However, earlier chunks are faster to generate due to an initially empty KV cache that gradually fills, so this average does not represent steady-state latency at long contexts. 
Occasionally, baselines provide ambiguous latency numbers without explanation; Causal Forcing++~\citep{zhao2026causal} reports 270 ms for its 1, 2, and 4 step models without explanation.

All of our profiling numbers include a breakdown of the specific operation, and are measured at steady-state with a full KV cache where applicable based on officially released configurations (e.g., KV=32760 for Self-Forcing and its follow-up works).

\section{Architecture and Training Details of Pixel Decoder}
Our pixel decoder is a causal ConvNeXt model operating directly on the pre-head features of the frozen two-step DiTs. For a 21-frame latent sequence, the DiT produces 32,760 tokens on a ($21\times30\times52$) spatiotemporal grid with channel width 1536. The head applies per-frame adaptive shift/scale modulation, a ($1\times1$) channel projection, and three ConvNeXt-style causal 3D residual blocks—two low-resolution blocks followed by one output block—using depthwise ($3\times3\times3$) convolutions, channel-wise layer normalization, and width-4 pointwise GELU expansions. All convolutional processing remains at the DiT token resolution; the selected configuration has no learned temporal or spatial upsampler, latent adapter, displacement branch, auxiliary feature fusion, or post-RGB refiner. A final per-token linear projection maps the 1536-dimensional features to ($4\times16\times16\times3$) RGB values, which are unpatchified into 84 frames at ($480\times832$) and aligned to
the VAE convention by dropping the first three frames, yielding 81 output frames. The decoder contains 63.87M parameters and is causal in time, so each output block
depends only on the current and preceding latent blocks. 

We train the pixel decoder using a paired spatial-temporal reconstruction objective. Let $\hat{y}_{\mathrm{pix}}$ denote the RGB video predicted by the ConvNeXt decoder, and let $y_{\mathrm{VAE}}$ denote the output obtained by decoding the same final latent $x_0$ with the frozen VAE. Both videos are
represented in the range $[-1,1]$. Our reconstruction objective is
  \begin{equation}
  \mathcal{L}_{\mathrm{rec}}=0.5\,\mathcal{L}_{1}+5.0\,\mathcal{L}_{\mathrm{MSE}}+0.05\,\mathcal{L}_{\mathrm{grad}}+0.05\,\mathcal{L}_{\mathrm{temp}}+0.50\,\mathcal{L}
  _{\mathrm{motion}}.
  \label{eq:pixel_reconstruction_supp}
  \end{equation}
The pixelwise terms are $\mathcal{L}_{1}=\operatorname{mean}(|\hat{y}_{\mathrm{pix}}-y_{\mathrm{VAE}}|)$ and $\mathcal{L}_{\mathrm{MSE}}=\operatorname{mean}((\hat{y}
_{\mathrm{pix}}-y_{\mathrm{VAE}})^2)$. To preserve spatial structure, we match horizontal and vertical finite differences:

\begin{equation}
  \mathcal{L}_{\mathrm{grad}}=\operatorname{mean}\!\left(|\Delta_h\hat{y}_{\mathrm{pix}}-\Delta_h y_{\mathrm{VAE}}|\right)+\operatorname{mean}\!\left(|\Delta_w\hat{y}
  _{\mathrm{pix}}-\Delta_w y_{\mathrm{VAE}}|\right).
  \label{eq:pixel_spatial_gradient}
  \end{equation}
  
To preserve temporal dynamics, we match consecutive-frame differences:
  \begin{equation}
  \mathcal{L}_{\mathrm{temp}}=\operatorname{mean}\!\left(|\Delta_t\hat{y}_{\mathrm{pix}}-\Delta_t y_{\mathrm{VAE}}|\right).
  \label{eq:pixel_temporal_difference}
  \end{equation}
Finally, we place greater reconstruction weight on regions that move in the VAE target:
\begin{equation}
\mathcal{L}_{\mathrm{motion}}=\operatorname{mean}_{t>1,p}\!\left[\left(1+\gamma\,\operatorname{mean}_{c}\left|\Delta_t y_{\mathrm{VAE},t,c,p}\right|
  \right)\operatorname{mean}_{c}\left|\hat{y}_{\mathrm{pix},t,c,p}-y_{\mathrm{VAE},t,c,p}\right|\right],\quad \gamma=10.
\label{eq:pixel_motion_weighted}
\end{equation}
Here, $\Delta_h$, $\Delta_w$, and $\Delta_t$ denote vertical, horizontal, and temporal finite-difference operators, respectively; $c$ indexes RGB channels; and $p$ indexes spatial locations. The multiplier $\gamma$ controls how strongly moving target regions are emphasized within $\mathcal{L}_{\mathrm{motion}}$, while the coefficient $0.50$ controls the overall contribution of this term to $\mathcal{L}_{\mathrm{rec}}$.

Using the \verb|vidprom_filtered_extended.txt| file provided in Self-Forcing's repo, we roll out the frozen DiT and VAE 150k times to get the input and output targets, running 4 optimization steps per rollout. We train on 32 B200 GPUs for about a week; we find that validation loss and metrics (PSNR and SSIM) continue to improve steadily though slowly.

We take the last checkpoint and train with the adversarial setup as described in the main paper (Sec.~\ref{sec:decoder}). 
We train on 8 B200 GPUs and run 5000 rollouts per GPU and 4 optimization steps per rollout. Training completes in a few hours.

\section{Architecture and Training Details of KV Refinement Network}

As described in the main paper, we first investigate whether KV refinement is necessary, given that the last diffusion sampling step writes a noisy KV cache for free. In Fig.~\ref{fig:kv_ablation}, we find that removing KV refinement, even after training the model to learn to use the noisy KV cache, leads to drift and artifacts. These experiments motivate our design.

Our KV refinement network is a cache-repair module that replaces the expensive DiT evaluation of the baselines. Each DiT layer maintains its own KV cache, partitioned into one slice for each generated latent frame. During every diffusion evaluation for frame $i$, the KV tensors for that frame overwrite its current cache slice, while the slices for earlier frames remain available as causal context. At the final denoising step, the DiT still receives a nonzero-noise latent, although it predicts the clean latent $\vz_0$; consequently, the KV written to the current slice correspond to the noisy input. The baseline therefore performs an additional, non-denoising DiT forward pass on $\vz_0$ with zero-noise time conditioning ($t=0$), overwriting the same slice with clean-context KV. 

Instead, our network consumes the noisy K/V tensors from the final denoising pass together with the predicted $\vz_0$, predicts corrected K/V tensors, and overwrites the same cache slice.
Subsequent latent frames therefore attend to the repaired context without requiring the additional full DiT forward pass.
A shared stem maps the 1,560 patch tokens per frame from 1,536 to 768 channels, then applies four lightweight 8-head spatial self-attention blocks with 4× GELU MLPs and timestep-conditioned adaptive layer normalization. Each of the DiT layers has a separate repair head that projects its noisy K/V, concatenates them with the shared per-token $\vz_0$ features, and predicts residual corrections. Keys are corrected in post-RoPE space and renormalized per attention head, while values receive a learned residual and per-channel scale. All residual outputs are zero-initialized so the adapter initially behaves like the no-prime baseline. 

During inference, frame 0 retains a real clean-prime anchor, while subsequent frames use the repaired K/V to overwrite their cache slots before the next frame is processed. The adapter is trained with the DiT and VAE frozen using teacher-forced targets from the real prime pass: the primary objective matches future attention outputs, supplemented by direct K/V reconstruction with approximately three times more weight on value than key errors, since we found that value corruption dominates the degradation.

\input{figures/kv_ablation.tex}

\section{Adaptation Details of Interactive Applications}
We use the official model and weights released by the authors. We follow the procedure described in Sec.~\ref{sec:method} to train both the pixel decoder and the KV refinement network. Neither component requires large-scale training data. We simply roll out the models to create target pixels and clean keys and values on-the-fly. 

LingBot requires four diffusion sampling steps per autoregressive generation, and generates four latent frames at once. RealWonder requires three sampling steps and generates three latent frames at once. Both are substantially more expensive than our setup in \name, which performs two sampling steps, and generates one latent frame at a time. Thus, even the 1.3B models require 8 GPUs to reach interactive latency. 
RealWonder is slower than LingBot at the same parameter count due to its more expensive processing of its action conditions. 
Distilling these models to fewer steps and fewer latent frames per generation while maintaining output quality and adherence to conditions is important future work.

%% file: tables/additional_profiling.tex
\begin{table}[h]                        
  \centering                                 
  \small                                      
  \begin{tabular}{lrrr}                        
  \toprule                                   
  Configuration & Resolution & B200 Latency (ms) & H100 Latency (ms) \\   
  \midrule                                    
  \multicolumn{4}{l}{\textbf{\name-S (1.3B) on 1 GPU}} \\   
  DiT evaluation  & 832×480 & 16 & 25\\
  DiT evaluation  & 1280×720 & 60 & 101\\
  KV refinement network  & 832×480 & 2 & 3\\
  KV refinement network  & 1280×720 & 5 & 7\\
  VAE & 832×480 & 147 & 175\\
  VAE & 1280×720 & 327 & 408\\
  Our pixel decoder & 832×480 & 1 & 1\\
  Our pixel decoder & 1280×720 & 2 & 3\\
  \midrule
  \multicolumn{4}{l}{\textbf{\name-L (14B) on 8 GPUs}} \\   
  DiT evaluation  & 832×480 & 20 & 30\\
  DiT evaluation  & 1280×720 & 54 & 67\\
  KV refinement network  & 832×480 & 3 & 5\\
  KV refinement network  & 1280×720 & 7  & 12\\
  Our pixel decoder & 832×480 & 1 & 1\\
  Our pixel decoder & 1280×720 & 1 & 1\\
  \bottomrule
  \end{tabular}
  \caption{
  \textbf{Additional latency profiling at 720p resolution and on H100 GPUs}. \name-S runs in 36 ms at 480p and in 127 ms at 720p on a B200 GPU. \name-L runs in 45 ms at 480p and in 116 ms at 720p on 8 B200 GPUs. Savings from our lightweight network replacements are more noticeable at 720p. H100s are approximately 25\% to 50\% slower depending on the configuration.}
  %Distributing the VAE across 8 GPUs is non-trivial; we leave it for future work.}
  \label{tab:lingbot_latency}
  \end{table}

%% file: figures/kv_ablation.tex
\begin{figure}
    \centering
    \includegraphics[width=\linewidth]{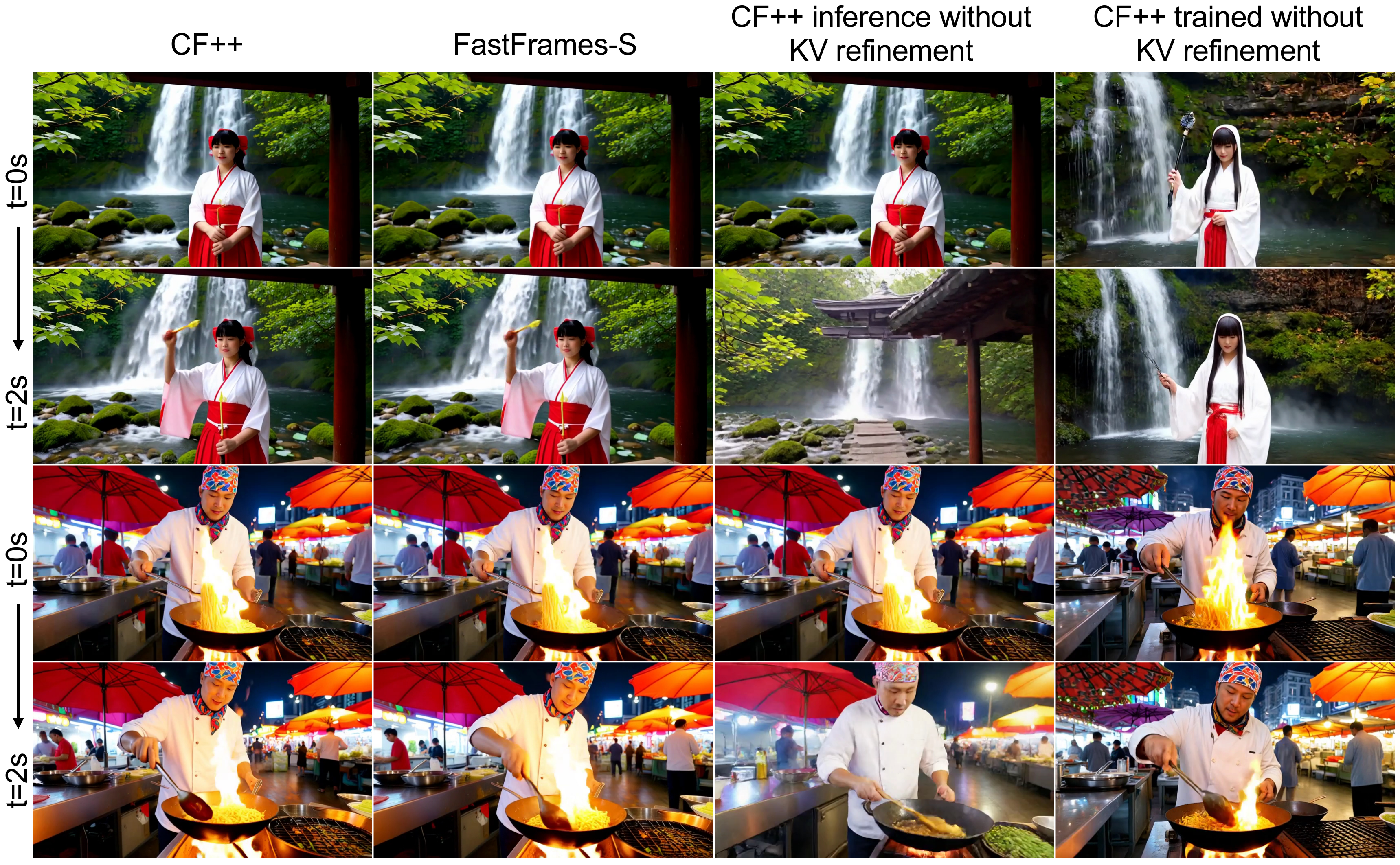}
    \caption{
    \textbf{KV refinement ablation.}
    Autoregressive baselines refine the KV cache using an additional DiT evaluation, which is costly but effective. Removing that refinement naively (third column) substantially degrades the scene's consistency; the objects in the scenes change over time. Even training without the refinement (fourth column) does not fix the problem. Either the subject changes over time (the face changes in the second example), or noticeable artifacts appear (clothing in the first example, fingers in the second example). Our KV  refinement network (second column) replaces the DiT evaluation and performs equally while being an order of magnitude faster. 
    }
    \label{fig:kv_ablation}
\end{figure}

%% file: main.bbl
\begin{thebibliography}{88}
\providecommand{\natexlab}[1]{#1}
\providecommand{\url}[1]{\texttt{#1}}
\expandafter\ifx\csname urlstyle\endcsname\relax
  \providecommand{\doi}[1]{doi: #1}\else
  \providecommand{\doi}{doi: \begingroup \urlstyle{rm}\Url}\fi

\bibitem[Annett et~al.(2014)Annett, Ng, Dietz, Bischof, and Gupta]{annett2014how}
Michelle Annett, Albert Ng, Paul Dietz, Walter~F. Bischof, and Anoop Gupta.
\newblock How low should we go? {Understanding} the perception of latency while inking.
\newblock In \emph{Graphics Interface}, 2014.

\bibitem[Ansel et~al.(2024)Ansel, Yang, He, Gimelshein, Jain, Voznesensky, Bao, Bell, Berard, Burovski, Chauhan, Chourdia, Constable, Desmaison, DeVito, Ellison, Feng, Gong, Gschwind, Hirsh, Huang, Kalambarkar, Kirsch, Lazos, Lezcano, Liang, Liang, Lu, Luk, Maher, Pan, Puhrsch, Reso, Saroufim, Siraichi, Suk, Suo, Tillet, Wang, Wang, Wen, Zhang, Zhao, Zhou, Zou, Mathews, Chanan, Wu, and Chintala]{ansel2024pytorch2}
Jason Ansel, Edward Yang, Horace He, Natalia Gimelshein, Animesh Jain, Michael Voznesensky, Bin Bao, Peter Bell, David Berard, Evgeni Burovski, Geeta Chauhan, Anjali Chourdia, Will Constable, Alban Desmaison, Zachary DeVito, Elias Ellison, Will Feng, Jiong Gong, Michael Gschwind, Brian Hirsh, Sherlock Huang, Kshiteej Kalambarkar, Laurent Kirsch, Michael Lazos, Mario Lezcano, Yanbo Liang, Jason Liang, Yinghai Lu, CK~Luk, Bert Maher, Yunjie Pan, Christian Puhrsch, Matthias Reso, Mark Saroufim, Marcos~Yukio Siraichi, Helen Suk, Michael Suo, Phil Tillet, Eikan Wang, Xiaodong Wang, William Wen, Shunting Zhang, Xu~Zhao, Keren Zhou, Richard Zou, Ajit Mathews, Gregory Chanan, Peng Wu, and Soumith Chintala.
\newblock {PyTorch 2}: Faster machine learning through dynamic {Python} bytecode transformation and graph compilation.
\newblock In \emph{ASPLOS}, 2024.

\bibitem[Bandyopadhyay et~al.(2026)Bandyopadhyay, Pinnaparaju, Entezari, Scott, Song, and Jampani]{bandyopadhyay2024blockcascading}
Hmrishav Bandyopadhyay, Nikhil Pinnaparaju, Rahim Entezari, Jim Scott, Yi-Zhe Song, and Varun Jampani.
\newblock {Block Cascading}: Training free acceleration of block-causal video models.
\newblock In \emph{CVPR Findings}, 2026.

\bibitem[{Black Forest Labs}(2026)]{blackforestlabs2026flux3}
{Black Forest Labs}.
\newblock {FLUX 3} -- real world models: Towards multimodal flow models as the backbone of visual intelligence, July 2026.
\newblock URL \url{https://bfl.ai/blog/flux-3}.
\newblock Model release.

\bibitem[Blattmann et~al.(2023)Blattmann, Dockhorn, Kulal, Mendelevitch, Kilian, Lorenz, Levi, English, Voleti, Letts, Jampani, and Rombach]{blattmann2023SVD}
Andreas Blattmann, Tim Dockhorn, Sumith Kulal, Daniel Mendelevitch, Maciej Kilian, Dominik Lorenz, Yam Levi, Zion English, Vikram Voleti, Adam Letts, Varun Jampani, and Robin Rombach.
\newblock {Stable Video Diffusion}: Scaling latent video diffusion models to large datasets.
\newblock \emph{arXiv:2311.15127}, 2023.

\bibitem[Boer~Bohan(2023)]{bohan2023taesd}
Ollin Boer~Bohan.
\newblock {TAESD}: Tiny autoencoder for {Stable Diffusion}.
\newblock \url{https://github.com/madebyollin/taesd}, 2023.

\bibitem[Boer~Bohan(2025)]{BoerBohan2025TAEHV}
Ollin Boer~Bohan.
\newblock {TAEHV}: Tiny autoencoder for {Hunyuan Video}.
\newblock \url{https://github.com/madebyollin/taehv}, 2025.

\bibitem[Buzovkin \& Shilov(2025)Buzovkin and Shilov]{buzovkin2024fastdecoders}
Alexey Buzovkin and Evgeny Shilov.
\newblock Toward lightweight and fast decoders for diffusion models in image and video generation.
\newblock \emph{arXiv:2503.04871}, 2025.

\bibitem[Chen et~al.(2025)Chen, Lin, Yang, Lin, Zhu, Fan, Zhang, Chen, Chen, Ma, Xiong, Wang, Pang, Kang, Xu, Jin, Liang, Song, Zhao, Xu, Qiu, Li, Fei, Li, and Zhou]{chen2025skyreelsv2infinitelengthfilmgenerative}
Guibin Chen, Dixuan Lin, Jiangping Yang, Chunze Lin, Junchen Zhu, Mingyuan Fan, Hao Zhang, Sheng Chen, Zheng Chen, Chengcheng Ma, Weiming Xiong, Wei Wang, Nuo Pang, Kang Kang, Zhiheng Xu, Yuzhe Jin, Yupeng Liang, Yubing Song, Peng Zhao, Boyuan Xu, Di~Qiu, Debang Li, Zhengcong Fei, Yang Li, and Yahui Zhou.
\newblock {SkyReels-V2}: Infinite-length film generative model.
\newblock \emph{arXiv:2504.13074}, 2025.

\bibitem[Chen et~al.(2026{\natexlab{a}})Chen, Zhao, Yu, Chu, Chen, Yang, Wang, Pan, Zhou, Ling, Liu, Yi, Zhang, Li, Chen, Cai, Fidler, Luo, Han, and Xie]{chen2025sanavideo}
Junsong Chen, Yuyang Zhao, Jincheng Yu, Ruihang Chu, Junyu Chen, Shuai Yang, Xianbang Wang, Yicheng Pan, Daquan Zhou, Huan Ling, Haozhe Liu, Hongwei Yi, Hao Zhang, Muyang Li, Yukang Chen, Han Cai, Sanja Fidler, Ping Luo, Song Han, and Enze Xie.
\newblock {SANA-Video}: Efficient video generation with block linear diffusion transformer.
\newblock In \emph{ICLR}, 2026{\natexlab{a}}.

\bibitem[Chen et~al.(2026{\natexlab{b}})Chen, Wang, Huang, Yang, Zhang, Xiao, Chu, Mao, Hu, Liu, Zhao, Mao, Chen, Xie, Qi, and Han]{chen2026longlive}
Yukang Chen, Luozhou Wang, Wei Huang, Shuai Yang, Bohan Zhang, Yicheng Xiao, Ruihang Chu, Weian Mao, Qixin Hu, Shaoteng Liu, Yuyang Zhao, Huizi Mao, Ying-Cong Chen, Enze Xie, Xiaojuan Qi, and Song Han.
\newblock {LongLive-2.0}: An {NVFP4} parallel infrastructure for long video generation.
\newblock \emph{arXiv:2605.18739}, 2026{\natexlab{b}}.

\bibitem[Chou et~al.(2025)Chou, Zhang, Bi, Tan, Xu, Luan, Hariharan, and Snavely]{chou2024kfcw}
Gene Chou, Kai Zhang, Sai Bi, Hao Tan, Zexiang Xu, Fujun Luan, Bharath Hariharan, and Noah Snavely.
\newblock Generating {3D}-consistent videos from unposed internet photos.
\newblock In \emph{CVPR}, 2025.

\bibitem[Chou et~al.(2026)Chou, Herrmann, Genova, Deng, Peng, Hariharan, Zhang, Snavely, and Henzler]{chou2026cityrag}
Gene Chou, Charles Herrmann, Kyle Genova, Boyang Deng, Songyou Peng, Bharath Hariharan, Jason~Y. Zhang, Noah Snavely, and Philipp Henzler.
\newblock {CityRAG}: Stepping into a city via spatially-grounded video generation.
\newblock In \emph{NeurIPS}, 2026.

\bibitem[Cui et~al.(2026)Cui, Wu, Li, Yang, Li, Wang, Bai, Ban, and Hsieh]{cui2025selfforcingpp}
Justin Cui, Jie Wu, Ming Li, Tao Yang, Xiaojie Li, Rui Wang, Andrew Bai, Yuanhao Ban, and Cho-Jui Hsieh.
\newblock {Self-Forcing++}: Towards minute-scale high-quality video generation.
\newblock In \emph{ICLR}, 2026.

\bibitem[Dao(2024)]{dao2023flashattention2}
Tri Dao.
\newblock {FlashAttention-2}: Faster attention with better parallelism and work partitioning.
\newblock In \emph{ICLR}, 2024.

\bibitem[Deber et~al.(2015)Deber, Jota, Forlines, and Wigdor]{deber2015how}
Jonathan Deber, Ricardo Jota, Clifton Forlines, and Daniel Wigdor.
\newblock How much faster is fast enough? {User} perception of latency \& latency improvements in direct and indirect touch.
\newblock In \emph{CHI}, 2015.

\bibitem[Deng et~al.(2025)Deng, Pan, Diao, Luo, Cui, Lu, Shan, Qi, and Wang]{deng2024nova}
Haoge Deng, Ting Pan, Haiwen Diao, Zhengxiong Luo, Yufeng Cui, Huchuan Lu, Shiguang Shan, Yonggang Qi, and Xinlong Wang.
\newblock Autoregressive video generation without vector quantization.
\newblock In \emph{ICLR}, 2025.

\bibitem[Deng et~al.(2026)Deng, Yin, Guo, Wang, Fang, Yuan, Yang, Wang, Liu, Huang, and Ma]{deng2025magref}
Yufan Deng, Yuanyang Yin, Xun Guo, Yizhi Wang, Jacob~Zhiyuan Fang, Shenghai Yuan, Yiding Yang, Angtian Wang, Bo~Liu, Haibin Huang, and Chongyang Ma.
\newblock {MAGREF}: Masked guidance for any-reference video generation with subject disentanglement.
\newblock In \emph{ICLR}, 2026.

\bibitem[Fang et~al.(2025{\natexlab{a}})Fang, Li, Ma, and Wang]{fang2024tinyfusion}
Gongfan Fang, Kunjun Li, Xinyin Ma, and Xinchao Wang.
\newblock {TinyFusion}: Diffusion transformers learned shallow.
\newblock In \emph{CVPR}, 2025{\natexlab{a}}.

\bibitem[Fang et~al.(2024)Fang, Pan, Sun, Li, and Wang]{fang2024xdit}
Jiarui Fang, Jinzhe Pan, Xibo Sun, Aoyu Li, and Jiannan Wang.
\newblock {xDiT}: An inference engine for diffusion transformers ({DiTs}) with massive parallelism.
\newblock \emph{arXiv:2411.01738}, 2024.

\bibitem[Fang et~al.(2025{\natexlab{b}})Fang, Pan, Li, Sun, and Wang]{fang2024pipefusion}
Jiarui Fang, Jinzhe Pan, Aoyu Li, Xibo Sun, and Jiannan Wang.
\newblock {PipeFusion}: Patch-level pipeline parallelism for diffusion transformers inference.
\newblock In \emph{NeurIPS}, 2025{\natexlab{b}}.

\bibitem[Feng et~al.(2026)Feng, Li, Yang, Xi, Li, Li, Zhang, Yang, Peng, Han, Agrawala, Keutzer, Kodaira, and Xu]{feng2025streamdiffusionv2}
Tianrui Feng, Zhi Li, Shuo Yang, Haocheng Xi, Muyang Li, Xiuyu Li, Lvmin Zhang, Keting Yang, Kelly Peng, Song Han, Maneesh Agrawala, Kurt Keutzer, Akio Kodaira, and Chenfeng Xu.
\newblock {StreamDiffusionV2}: A streaming system for dynamic and interactive video generation.
\newblock In \emph{MLSys}, 2026.

\bibitem[Forch et~al.(2017)Forch, Franke, Rauh, and Krems]{forch2017are}
Valentin Forch, Thomas Franke, Nadine Rauh, and Josef~F. Krems.
\newblock Are 100 ms fast enough? {C}haracterizing latency perception thresholds in mouse-based interaction.
\newblock In \emph{Engineering Psychology and Cognitive Ergonomics: Cognition and Design (HCI International)}, 2017.

\bibitem[Frans et~al.(2025)Frans, Hafner, Levine, and Abbeel]{frans2025shortcut}
Kevin Frans, Danijar Hafner, Sergey Levine, and Pieter Abbeel.
\newblock One step diffusion via shortcut models.
\newblock In \emph{ICLR}, 2025.

\bibitem[Gao et~al.(2026)Gao, Wang, Zhu, Chen, Liu, Bai, Wang, Yuan, Wang, Lu, Cheng, Zhang, Gao, Feng, Liu, Yao, Xu, Zhu, Shen, and Ouyang]{lingbot-world-v2}
Zelin Gao, Qiuyu Wang, Jiapeng Zhu, Jingye Chen, Zichen Liu, Qingyan Bai, Jiahao Wang, Yufeng Yuan, Hanlin Wang, Yichong Lu, Ka~Leong Cheng, Haojie Zhang, Jian Gao, Tianrui Feng, Yuzheng Liu, Yao Yao, Yinghao Xu, Xing Zhu, Yujun Shen, and Hao Ouyang.
\newblock Infinite worlds with versatile interactions.
\newblock \emph{arXiv:2607.07534}, 2026.

\bibitem[Geng et~al.(2025)Geng, Deng, Bai, Kolter, and He]{geng2026meanflow}
Zhengyang Geng, Mingyang Deng, Xingjian Bai, J.~Zico Kolter, and Kaiming He.
\newblock Mean flows for one-step generative modeling.
\newblock In \emph{NeurIPS}, 2025.

\bibitem[Ghosh et~al.(2025)Ghosh, Nayak, Panwar, and Basu]{ghosh2025pygraph}
Abhishek Ghosh, Ajay Nayak, Ashish Panwar, and Arkaprava Basu.
\newblock {PyGraph}: Robust compiler support for {CUDA} graphs in {PyTorch}.
\newblock \emph{arXiv:2503.19779}, 2025.

\bibitem[Goodfellow et~al.(2014)Goodfellow, Pouget-Abadie, Mirza, Xu, Warde-Farley, Ozair, Courville, and Bengio]{goodfellow2014generative}
Ian~J. Goodfellow, Jean Pouget-Abadie, Mehdi Mirza, Bing Xu, David Warde-Farley, Sherjil Ozair, Aaron Courville, and Yoshua Bengio.
\newblock Generative adversarial nets.
\newblock In \emph{NIPS}, 2014.

\bibitem[Guo et~al.(2026)Guo, Jia, Li, Li, Cai, Wang, Li, and Lu]{guo2026dummy}
Hang Guo, Zhaoyang Jia, Jiahao Li, Bin Li, Yuanhao Cai, Jiangshan Wang, Yawei Li, and Yan Lu.
\newblock Efficient autoregressive video diffusion with dummy head.
\newblock \emph{arXiv:2601.20499}, 2026.

\bibitem[HaCohen et~al.(2026)HaCohen, Brazowski, Chiprut, Bitterman, Kvochko, Berkowitz, Shalem, Lifschitz, Moshe, Porat, Richardson, Shiran, Chachy, Chetboun, Finkelson, Kupchick, Zabari, Guetta, Kotler, Bibi, Gordon, Panet, Benita, Armon, Kulikov, Inger, Shiftan, Melumian, and Farbman]{hacohen2026ltx2}
Yoav HaCohen, Benny Brazowski, Nisan Chiprut, Yaki Bitterman, Andrew Kvochko, Avishai Berkowitz, Daniel Shalem, Daphna Lifschitz, Dudu Moshe, Eitan Porat, Eitan Richardson, Guy Shiran, Itay Chachy, Jonathan Chetboun, Michael Finkelson, Michael Kupchick, Nir Zabari, Nitzan Guetta, Noa Kotler, Ofir Bibi, Ori Gordon, Poriya Panet, Roi Benita, Shahar Armon, Victor Kulikov, Yaron Inger, Yonatan Shiftan, Zeev Melumian, and Zeev Farbman.
\newblock {LTX-2}: Efficient joint audio-visual foundation model.
\newblock \emph{arXiv:2601.03233}, 2026.

\bibitem[He et~al.(2025)He, Xu, Guo, Wetzstein, Dai, Li, and Yang]{he2025cameractrl}
Hao He, Yinghao Xu, Yuwei Guo, Gordon Wetzstein, Bo~Dai, Hongsheng Li, and Ceyuan Yang.
\newblock {CameraCtrl}: Enabling camera control for video diffusion models.
\newblock In \emph{ICLR}, 2025.

\bibitem[Ho et~al.(2020)Ho, Jain, and Abbeel]{ho2020denoising}
Jonathan Ho, Ajay Jain, and Pieter Abbeel.
\newblock Denoising diffusion probabilistic models.
\newblock In \emph{NeurIPS}, 2020.

\bibitem[Ho et~al.(2022{\natexlab{a}})Ho, Chan, Saharia, Whang, Gao, Gritsenko, Kingma, Poole, Norouzi, Fleet, and Salimans]{ho2022imagenvideo}
Jonathan Ho, William Chan, Chitwan Saharia, Jay Whang, Ruiqi Gao, Alexey Gritsenko, Diederik~P. Kingma, Ben Poole, Mohammad Norouzi, David~J. Fleet, and Tim Salimans.
\newblock {Imagen Video}: High definition video generation with diffusion models.
\newblock \emph{arXiv:2210.02303}, 2022{\natexlab{a}}.

\bibitem[Ho et~al.(2022{\natexlab{b}})Ho, Salimans, Gritsenko, Chan, Norouzi, and Fleet]{ho2022videodiffusion}
Jonathan Ho, Tim Salimans, Alexey Gritsenko, William Chan, Mohammad Norouzi, and David~J. Fleet.
\newblock Video diffusion models.
\newblock In \emph{NeurIPS}, 2022{\natexlab{b}}.

\bibitem[Huang et~al.(2025)Huang, Li, He, Zhou, and Shechtman]{huang2026selfforcing}
Xun Huang, Zhengqi Li, Guande He, Mingyuan Zhou, and Eli Shechtman.
\newblock {Self Forcing}: Bridging the train-test gap in autoregressive video diffusion.
\newblock In \emph{NeurIPS}, 2025.

\bibitem[Huang et~al.(2024)Huang, He, Yu, Zhang, Si, Jiang, Zhang, Wu, Jin, Chanpaisit, Wang, Chen, Wang, Lin, Qiao, and Liu]{huang2023vbench}
Ziqi Huang, Yinan He, Jiashuo Yu, Fan Zhang, Chenyang Si, Yuming Jiang, Yuanhan Zhang, Tianxing Wu, Qingyang Jin, Nattapol Chanpaisit, Yaohui Wang, Xinyuan Chen, Limin Wang, Dahua Lin, Yu~Qiao, and Ziwei Liu.
\newblock {VBench}: Comprehensive benchmark suite for video generative models.
\newblock In \emph{CVPR}, 2024.

\bibitem[Jacobs et~al.(2024)Jacobs, Tanaka, Zhang, Zhang, Aminabadi, Song, Rajbhandari, and He]{jacobs2023deepspeedulysses}
Sam~Ade Jacobs, Masahiro Tanaka, Chengming Zhang, Minjia Zhang, Reza~Yazdani Aminabadi, Shuaiwen~Leon Song, Samyam Rajbhandari, and Yuxiong He.
\newblock System optimizations for enabling training of extreme long sequence transformer models.
\newblock In \emph{PODC}, 2024.

\bibitem[Ji et~al.(2026)Ji, Zhong, Zhang, Yang, Jin, Qin, Luo, Mao, Liu, and Li]{ji2026forcingkv}
Yicheng Ji, Zhizhou Zhong, Jun Zhang, Qin Yang, XiTai Jin, Ying Qin, Wenhan Luo, Shuiyang Mao, Wei Liu, and Huan Li.
\newblock {Forcing-KV}: Hybrid {KV} cache compression for efficient autoregressive video diffusion models.
\newblock \emph{arXiv:2605.09681}, 2026.

\bibitem[Jiang et~al.(2026)Jiang, Sun, Wang, Zhu, Wang, Zhang, Liu, Wang, Zheng, Sun, Ni, Pan, Sun, Yu, Ge, Du, Fan, Sun, Liu, Yongchang, Zhu, Wang, Ying, Xuan, Yang, Liu, Gao, Xu, Sui, Yang, Lai, Liu, Liu, Zhou, Peng, Cao, Sheng, Cai, Lu, Xu, and Guo]{jiang2026abotworld0}
Fan Jiang, Zhaoxu Sun, Mengchao Wang, Ziyu Zhu, Chiyu Wang, Yunpeng Zhang, Wenlin Liu, Yun Wang, Xue Zheng, Rui Sun, Junfeng Ni, Hongyu Pan, Zhongxu Sun, Fei Yu, Zengye Ge, Mengmeng Du, Nianfei Fan, Mingchao Sun, Yu~Liu, Yongchang, Yanqing Zhu, Jiahang Wang, Ning Ying, Yuze Xuan, Di~Yang, Zhicheng Liu, Zhe Gao, Tingbing Xu, Jiacheng Sui, Wenjin Yang, Junnan Lai, Shufeng Liu, Yuan Liu, Zheng Zhou, Yingliang Peng, Dawei Cao, Kaifeng Sheng, Yuxiang Cai, Fei Lu, Mu~Xu, and Ning Guo.
\newblock {ABot-World-0}: Infinite interactive world rollout on a single desktop {GPU}.
\newblock \emph{arXiv:2607.19191}, 2026.

\bibitem[Jin et~al.(2025)Jin, Sun, Li, Xu, Xu, Jiang, Zhuang, Huang, Song, Mu, and Lin]{jin2024pyramidal}
Yang Jin, Zhicheng Sun, Ningyuan Li, Kun Xu, Kun Xu, Hao Jiang, Nan Zhuang, Quzhe Huang, Yang Song, Yadong Mu, and Zhouchen Lin.
\newblock Pyramidal flow matching for efficient video generative modeling.
\newblock In \emph{ICLR}, 2025.

\bibitem[Jolicoeur-Martineau(2019)]{jolicoeur2018relativistic}
Alexia Jolicoeur-Martineau.
\newblock The relativistic discriminator: a key element missing from standard {GAN}.
\newblock In \emph{ICLR}, 2019.

\bibitem[Kingma \& Welling(2014)Kingma and Welling]{kingma2014autoencoding}
Diederik~P. Kingma and Max Welling.
\newblock Auto-encoding variational {Bayes}.
\newblock In \emph{ICLR}, 2014.

\bibitem[Kodaira et~al.(2025)Kodaira, Xu, Hazama, Yoshimoto, Ohno, Mitsuhori, Sugano, Cho, Liu, Tomizuka, and Keutzer]{kodaira2023streamdiffusion}
Akio Kodaira, Chenfeng Xu, Toshiki Hazama, Takanori Yoshimoto, Kohei Ohno, Shogo Mitsuhori, Soichi Sugano, Hanying Cho, Zhijian Liu, Masayoshi Tomizuka, and Kurt Keutzer.
\newblock {StreamDiffusion}: {A} pipeline-level solution for real-time interactive generation.
\newblock In \emph{ICCV}, 2025.

\bibitem[Li et~al.(2026)Li, Shah, and Shang]{li2026attend}
Kunyang Li, Mubarak Shah, and Yuzhang Shang.
\newblock Attend locally, remember linearly: Linear attention as cross-frame memory for autoregressive video diffusion.
\newblock In \emph{NeurIPS}, 2026.

\bibitem[Li et~al.(2024)Li, Cai, Cao, Zhang, Cai, Bai, Jia, Liu, Li, and Han]{li2024distrifusion}
Muyang Li, Tianle Cai, Jiaxin Cao, Qinsheng Zhang, Han Cai, Junjie Bai, Yangqing Jia, Ming-Yu Liu, Kai Li, and Song Han.
\newblock {DistriFusion}: Distributed parallel inference for high-resolution diffusion models.
\newblock In \emph{CVPR}, 2024.

\bibitem[Li et~al.(2025{\natexlab{a}})Li, Lin, Zhang, Cai, Li, Guo, Xie, Meng, Zhu, and Han]{li2024svdquant}
Muyang Li, Yujun Lin, Zhekai Zhang, Tianle Cai, Xiuyu Li, Junxian Guo, Enze Xie, Chenlin Meng, Jun-Yan Zhu, and Song Han.
\newblock {SVDQuant}: Absorbing outliers by low-rank components for 4-bit diffusion models.
\newblock In \emph{ICLR}, 2025{\natexlab{a}}.

\bibitem[Li \& He(2026)Li and He]{li2025jit}
Tianhong Li and Kaiming He.
\newblock Back to basics: Let denoising generative models denoise.
\newblock In \emph{CVPR}, 2026.

\bibitem[Li et~al.(2025{\natexlab{b}})Li, Yu, Liu, Yang, Herrmann, Wetzstein, and Wu]{li2025wonderplay}
Zizhang Li, Hong-Xing Yu, Wei Liu, Yin Yang, Charles Herrmann, Gordon Wetzstein, and Jiajun Wu.
\newblock {WonderPlay}: Dynamic {3D} scene generation from a single image and actions.
\newblock In \emph{ICCV}, 2025{\natexlab{b}}.

\bibitem[Lin et~al.(2026)Lin, Yang, Lin, Chen, and Fan]{afm}
Shanchuan Lin, Ceyuan Yang, Zhijie Lin, Hao Chen, and Haoqi Fan.
\newblock Adversarial flow models.
\newblock In \emph{ICML}, 2026.

\bibitem[Liu et~al.(2026{\natexlab{a}})Liu, Gao, Liu, Du, Li, Wu, Jin, Cao, Zhang, Hoi, and Li]{liu2025decoupled}
Dongyang Liu, Peng Gao, David Liu, Ruoyi Du, Zhen Li, Qilong Wu, Xin Jin, Sihan Cao, Shifeng Zhang, Steven Hoi, and Hongsheng Li.
\newblock Decoupled {DMD:} {CFG} augmentation as the spear, distribution matching as the shield.
\newblock In \emph{ICLR}, 2026{\natexlab{a}}.

\bibitem[Liu et~al.(2026{\natexlab{b}})Liu, Liu, Mei, Wen, Yang, and Liu]{liu2026diagonal}
Jinxiu Liu, Xuanming Liu, Kangfu Mei, Yandong Wen, Ming-Hsuan Yang, and Weiyang Liu.
\newblock Streaming autoregressive video generation via diagonal distillation.
\newblock In \emph{ICLR}, 2026{\natexlab{b}}.

\bibitem[Liu et~al.(2026{\natexlab{c}})Liu, Chen, Li, Wang, Yu, and Wu]{liu2026realwonder}
Wei Liu, Ziyu Chen, Zizhang Li, Yue Wang, Hong{-}Xing Yu, and Jiajun Wu.
\newblock {RealWonder}: Real-time physical action-conditioned video generation.
\newblock \emph{arXiv:2603.05449}, 2026{\natexlab{c}}.

\bibitem[Liu et~al.(2022)Liu, Mao, Wu, Feichtenhofer, Darrell, and Xie]{liu2022convnet}
Zhuang Liu, Hanzi Mao, Chao-Yuan Wu, Christoph Feichtenhofer, Trevor Darrell, and Saining Xie.
\newblock A {ConvNet} for the 2020s.
\newblock In \emph{CVPR}, 2022.

\bibitem[Lu et~al.(2022)Lu, Zhou, Bao, Chen, Li, and Zhu]{lu2022dpm}
Cheng Lu, Yuhao Zhou, Fan Bao, Jianfei Chen, Chongxuan Li, and Jun Zhu.
\newblock {DPM-Solver}: {A} fast {ODE} solver for diffusion probabilistic model sampling in around 10 steps.
\newblock In \emph{NeurIPS}, 2022.

\bibitem[Lu et~al.(2025)Lu, Zhou, Bao, Chen, Li, and Zhu]{lu2025dpm}
Cheng Lu, Yuhao Zhou, Fan Bao, Jianfei Chen, Chongxuan Li, and Jun Zhu.
\newblock {DPM-Solver++}: Fast solver for guided sampling of diffusion probabilistic models.
\newblock \emph{Machine Intelligence Research}, 22\penalty0 (4), 2025.

\bibitem[Miller(1968)]{miller1968response}
Robert~B. Miller.
\newblock Response time in man-computer conversational transactions.
\newblock In \emph{AFIPS Fall Joint Computer Conference}, 1968.

\bibitem[Millon(2025)]{krea_realtime_14b}
Erwann Millon.
\newblock {Krea Realtime 14B}: Real-time video generation, 2025.
\newblock URL \url{https://github.com/krea-ai/realtime-video}.

\bibitem[{MiniMax}(2026)]{minimax2026h3}
{MiniMax}.
\newblock {MiniMax H3}: An open model breaking the boundaries between tasks and modalities, July 2026.
\newblock URL \url{https://www.minimax.io/blog/minimax-h3}.
\newblock Model release.

\bibitem[{NVIDIA}(2026)]{nvidia2026cosmos3}
{NVIDIA}.
\newblock Cosmos 3: Omnimodal world models for physical {AI}.
\newblock \emph{arXiv:2606.02800}, 2026.

\bibitem[{NVIDIA Corporation}(2026)]{nvidia2026cudnnattention}
{NVIDIA Corporation}.
\newblock {NVIDIA cuDNN}: Scaled dot product attention.
\newblock \url{https://docs.nvidia.com/deeplearning/cudnn/latest/operations/Attention.html}, 2026.
\newblock cuDNN documentation, accessed September 2, 2026.

\bibitem[Pan et~al.(2026)Pan, Tian, Zhang, Kong, Xiong, Long, Zhang, Qiu, Wang, Lv, Tu, Wu, Bo, Tang, and Zhong]{pan2026omniweaving}
Kaihang Pan, Qi~Tian, Jianwei Zhang, Weijie Kong, Jiangfeng Xiong, Yanxin Long, Shixue Zhang, Haiyi Qiu, Tan Wang, Zheqi Lv, Shuyuan Tu, Yue Wu, Liefeng Bo, Siliang Tang, and Zhao Zhong.
\newblock {OmniWeaving}: Towards unified video generation with free-form composition and reasoning.
\newblock \emph{arXiv:2603.24458}, 2026.

\bibitem[{Robbyant Team} et~al.(2026){Robbyant Team}, Gao, Wang, Zeng, Zhu, Cheng, Li, Wang, Xu, Ma, Chen, Liu, Cheng, Yao, Zhu, Meng, Zheng, Bai, Chen, Shen, Yu, Zhu, Shen, and Ouyang]{lingbot-world}
{Robbyant Team}, Zelin Gao, Qiuyu Wang, Yanhong Zeng, Jiapeng Zhu, Ka~Leong Cheng, Yixuan Li, Hanlin Wang, Yinghao Xu, Shuailei Ma, Yihang Chen, Jie Liu, Yansong Cheng, Yao Yao, Jiayi Zhu, Yihao Meng, Kecheng Zheng, Qingyan Bai, Jingye Chen, Zehong Shen, Yue Yu, Xing Zhu, Yujun Shen, and Hao Ouyang.
\newblock Advancing open-source world models.
\newblock \emph{arXiv:2601.20540}, 2026.

\bibitem[{Seedance Team} et~al.(2026)]{seedance2026seedance2}
{Seedance Team} et~al.
\newblock Seedance 2.0: Advancing video generation for world complexity.
\newblock \emph{arXiv:2604.14148}, 2026.

\bibitem[Seo et~al.(2026)Seo, Choi, Kwon, Choi, Jin, Lee, Kim, Lee, Gu, Han, Yun, Kim, and Kim]{seo2026grounding}
Junyoung Seo, Hyunwook Choi, Minkyung Kwon, Jinhyeok Choi, Siyoon Jin, Gayoung Lee, Junho Kim, JoungBin Lee, Geonmo Gu, Dongyoon Han, Sangdoo Yun, Seungryong Kim, and Jin-Hwa Kim.
\newblock {Seoul World Model}: Grounding world simulation models in a real-world metropolis.
\newblock In \emph{ECCV}, 2026.

\bibitem[Shaul et~al.(2026)Shaul, Liu, Vahdat, and Berner]{shaul2026parallel}
Neta Shaul, Chao Liu, Arash Vahdat, and Julius Berner.
\newblock Parallel decoding distillation for fast image and video generation.
\newblock \emph{arXiv:2607.26004}, 2026.

\bibitem[Shin et~al.(2026)Shin, Li, Zhang, Zhu, Park, Shechtman, and Huang]{shin2026motionstream}
Joonghyuk Shin, Zhengqi Li, Richard Zhang, Jun-Yan Zhu, Jaesik Park, Eli Shechtman, and Xun Huang.
\newblock {MotionStream}: Real-time video generation with interactive motion controls.
\newblock In \emph{ICLR}, 2026.

\bibitem[Sim{\'e}oni et~al.(2026)Sim{\'e}oni, Vo, Seitzer, Baldassarre, Oquab, Jose, Khalidov, Szafraniec, Yi, Ramamonjisoa, Massa, Haziza, Wehrstedt, Wang, Darcet, Moutakanni, Sentana, Roberts, Vedaldi, Tolan, Brandt, Couprie, Mairal, J{\'e}gou, Labatut, and Bojanowski]{simeoni2025dinov3}
Oriane Sim{\'e}oni, Huy~V. Vo, Maximilian Seitzer, Federico Baldassarre, Maxime Oquab, Cijo Jose, Vasil Khalidov, Marc Szafraniec, Seungeun Yi, Micha{\"e}l Ramamonjisoa, Francisco Massa, Daniel Haziza, Luca Wehrstedt, Jianyuan Wang, Timoth{\'e}e Darcet, Th{\'e}o Moutakanni, Leonel Sentana, Claire Roberts, Andrea Vedaldi, Jamie Tolan, John Brandt, Camille Couprie, Julien Mairal, Herv{\'e} J{\'e}gou, Patrick Labatut, and Piotr Bojanowski.
\newblock {DINOv3}.
\newblock \emph{TMLR}, 2026.

\bibitem[Singer et~al.(2023)Singer, Polyak, Hayes, Yin, An, Zhang, Hu, Yang, Ashual, Gafni, Parikh, Gupta, and Taigman]{singer2022makeavideo}
Uriel Singer, Adam Polyak, Thomas Hayes, Xi~Yin, Jie An, Songyang Zhang, Qiyuan Hu, Harry Yang, Oron Ashual, Oran Gafni, Devi Parikh, Sonal Gupta, and Yaniv Taigman.
\newblock {Make-A-Video}: Text-to-video generation without text-video data.
\newblock In \emph{ICLR}, 2023.

\bibitem[Song et~al.(2021)Song, Meng, and Ermon]{song2020ddim}
Jiaming Song, Chenlin Meng, and Stefano Ermon.
\newblock Denoising diffusion implicit models.
\newblock In \emph{ICLR}, 2021.

\bibitem[Su et~al.(2026)Su, Zhang, Yuan, Duanmu, Chen, and Zhu]{su2026sixbitdiffusion}
Rundong Su, Jintao Zhang, Zhihang Yuan, Haojie Duanmu, Jianfei Chen, and Jun Zhu.
\newblock {6Bit-Diffusion}: Inference-time mixed-precision quantization for video diffusion models.
\newblock \emph{arXiv:2603.18742}, 2026.

\bibitem[Teng et~al.(2025)Teng, Jia, Sun, Li, Li, Tang, Han, Zhang, Zhang, Luo, Kang, Sun, Cao, Huang, Lin, Fang, Tao, Zhang, Wang, Liu, Shi, Su, Sun, Pan, Wang, Sheng, Cui, Hu, Yan, Yin, Zhang, Liu, Yin, Yang, Song, Hu, Zhang, and Li]{ai2025magi1autoregressivevideogeneration}
Hansi Teng, Hongyu Jia, Lei Sun, Lingzhi Li, Maolin Li, Mingqiu Tang, Shuai Han, Tianning Zhang, W.~Q. Zhang, Weifeng Luo, Xiaoyang Kang, Yuchen Sun, Yue Cao, Yunpeng Huang, Yutong Lin, Yuxin Fang, Zewei Tao, Zheng Zhang, Zhongshu Wang, Zixun Liu, Dai Shi, Guoli Su, Hanwen Sun, Hong Pan, Jie Wang, Jiexin Sheng, Min Cui, Min Hu, Ming Yan, Shucheng Yin, Siran Zhang, Tingting Liu, Xianping Yin, Xiaoyu Yang, Xin Song, Xuan Hu, Yankai Zhang, and Yuqiao Li.
\newblock {MAGI-1}: Autoregressive video generation at scale.
\newblock \emph{arXiv:2505.13211}, 2025.

\bibitem[Wang et~al.(2025)Wang, Ai, Wen, Mao, Xie, Chen, Yu, Zhao, Yang, Zeng, Wang, Zhang, Zhou, Wang, Chen, Zhu, Zhao, Yan, Huang, Feng, Zhang, Li, Wu, Chu, Feng, Zhang, Sun, Fang, Wang, Gui, Weng, Shen, Lin, Wang, Wang, Zhou, Wang, Shen, Yu, Shi, Huang, Xu, Kou, Lv, Li, Liu, Wang, Zhang, Huang, Li, Wu, Liu, Pan, Zheng, Hong, Shi, Feng, Jiang, Han, Wu, and Liu]{wan2025wan}
Ang Wang, Baole Ai, Bin Wen, Chaojie Mao, Chen{-}Wei Xie, Di~Chen, Feiwu Yu, Haiming Zhao, Jianxiao Yang, Jianyuan Zeng, Jiayu Wang, Jingfeng Zhang, Jingren Zhou, Jinkai Wang, Jixuan Chen, Kai Zhu, Kang Zhao, Keyu Yan, Lianghua Huang, Mengyang Feng, Ningyi Zhang, Pandeng Li, Pingyu Wu, Ruihang Chu, Ruili Feng, Shiwei Zhang, Siyang Sun, Tao Fang, Tianxing Wang, Tianyi Gui, Tingyu Weng, Tong Shen, Wei Lin, Wei Wang, Wei Wang, Wenmeng Zhou, Wente Wang, Wenting Shen, Wenyuan Yu, Xianzhong Shi, Xiaoming Huang, Xin Xu, Yan Kou, Yangyu Lv, Yifei Li, Yijing Liu, Yiming Wang, Yingya Zhang, Yitong Huang, Yong Li, You Wu, Yu~Liu, Yulin Pan, Yun Zheng, Yuntao Hong, Yupeng Shi, Yutong Feng, Zeyinzi Jiang, Zhen Han, Zhi{-}Fan Wu, and Ziyu Liu.
\newblock Wan: Open and advanced large-scale video generative models.
\newblock \emph{arXiv:2503.20314}, 2025.

\bibitem[Wang et~al.(2024)Wang, Yuan, Wang, Li, Chen, Xia, Luo, and Shan]{wang2024motionctrl}
Zhouxia Wang, Ziyang Yuan, Xintao Wang, Yaowei Li, Tianshui Chen, Menghan Xia, Ping Luo, and Ying Shan.
\newblock {MotionCtrl}: A unified and flexible motion controller for video generation.
\newblock In \emph{SIGGRAPH}, 2024.

\bibitem[Wu et~al.(2025{\natexlab{a}})Wu, Zou, Li, Huang, Yang, Tan, Peng, Wu, Xiong, Jiang, et~al.]{wu2025hunyuanvideo15}
Bing Wu, Chang Zou, Changlin Li, Duojun Huang, Fang Yang, Hao Tan, Jack Peng, Jianbing Wu, Jiangfeng Xiong, Jie Jiang, et~al.
\newblock {HunyuanVideo 1.5} technical report.
\newblock \emph{arXiv:2511.18870}, 2025{\natexlab{a}}.

\bibitem[Wu et~al.(2025{\natexlab{b}})Wu, Chen, Wang, and Xu]{wu2025individual}
Yiming Wu, Zhenghao Chen, Huan Wang, and Dong Xu.
\newblock Individual content and motion dynamics preserved pruning for video diffusion models.
\newblock In \emph{ACM MM}, 2025{\natexlab{b}}.

\bibitem[Xiao et~al.(2024)Xiao, Tian, Chen, Han, and Lewis]{xiao2024streaming}
Guangxuan Xiao, Yuandong Tian, Beidi Chen, Song Han, and Mike Lewis.
\newblock Efficient streaming language models with attention sinks.
\newblock In \emph{ICLR}, 2024.

\bibitem[Ye et~al.(2026)Ye, Ge, Zheng, Gao, Yu, Kurian, Indupuru, Tan, Zhu, Xiang, Malik, Lee, Liang, Ranawaka, Gu, Xu, Wang, Hu, Narayan, Bjorck, Wang, Kim, Niu, Zheng, Xie, Wu, Wang, Julian, Xu, Du, Chebotar, Reed, Kautz, Zhu, Fan, and Jang]{ye2026worldactionmodelszeroshot}
Seonghyeon Ye, Yunhao Ge, Kaiyuan Zheng, Shenyuan Gao, Sihyun Yu, George Kurian, Suneel Indupuru, You~Liang Tan, Chuning Zhu, Jiannan Xiang, Ayaan Malik, Kyungmin Lee, William Liang, Nadun Ranawaka, Jiasheng Gu, Yinzhen Xu, Guanzhi Wang, Fengyuan Hu, Avnish Narayan, Johan Bjorck, Jing Wang, Gwanghyun Kim, Dantong Niu, Ruijie Zheng, Yuqi Xie, Jimmy Wu, Qi~Wang, Ryan Julian, Danfei Xu, Yilun Du, Yevgen Chebotar, Scott Reed, Jan Kautz, Yuke Zhu, Linxi~``Jim'' Fan, and Joel Jang.
\newblock World action models are zero-shot policies.
\newblock In \emph{CoRL}, 2026.

\bibitem[Yin et~al.(2024{\natexlab{a}})Yin, Gharbi, Park, Zhang, Shechtman, Durand, and Freeman]{yin2024dmd2}
Tianwei Yin, Micha{\"e}l Gharbi, Taesung Park, Richard Zhang, Eli Shechtman, Fr{\'e}do Durand, and William~T. Freeman.
\newblock Improved distribution matching distillation for fast image synthesis.
\newblock In \emph{NeurIPS}, 2024{\natexlab{a}}.

\bibitem[Yin et~al.(2024{\natexlab{b}})Yin, Gharbi, Zhang, Shechtman, Durand, Freeman, and Park]{yin2024dmd}
Tianwei Yin, Micha{\"e}l Gharbi, Richard Zhang, Eli Shechtman, Fr{\'e}do Durand, William~T. Freeman, and Taesung Park.
\newblock One-step diffusion with distribution matching distillation.
\newblock In \emph{CVPR}, 2024{\natexlab{b}}.

\bibitem[Yin et~al.(2025)Yin, Zhang, Zhang, Freeman, Durand, Shechtman, and Huang]{yin2024causvid}
Tianwei Yin, Qiang Zhang, Richard Zhang, William~T. Freeman, Fr{\'{e}}do Durand, Eli Shechtman, and Xun Huang.
\newblock From slow bidirectional to fast autoregressive video diffusion models.
\newblock In \emph{CVPR}, 2025.

\bibitem[Yu et~al.(2025)Yu, Qin, Wang, Wan, Zhang, and Liu]{yu2025gamefactory}
Jiwen Yu, Yiran Qin, Xintao Wang, Pengfei Wan, Di~Zhang, and Xihui Liu.
\newblock {GameFactory}: Creating new games with generative interactive videos.
\newblock In \emph{ICCV}, 2025.

\bibitem[Yuan et~al.(2026)Yuan, Yin, Li, Huang, Yang, and Yuan]{helios}
Shenghai Yuan, Yuanyang Yin, Zongjian Li, Xinwei Huang, Xiao Yang, and Li~Yuan.
\newblock Helios: Real real-time long video generation model.
\newblock \emph{IEEE Transactions on Pattern Analysis and Machine Intelligence}, 2026.

\bibitem[Zhang et~al.(2025)Zhang, Chen, Su, Ding, Stoica, Liu, and Zhang]{zhang2025sliding}
Peiyuan Zhang, Yongqi Chen, Runlong Su, Hangliang Ding, Ion Stoica, Zhengzhong Liu, and Hao Zhang.
\newblock Fast video generation with sliding tile attention.
\newblock In \emph{ICML}, 2025.

\bibitem[Zhang et~al.(2018)Zhang, Isola, Efros, Shechtman, and Wang]{zhang2018perceptual}
Richard Zhang, Phillip Isola, Alexei~A. Efros, Eli Shechtman, and Oliver Wang.
\newblock The unreasonable effectiveness of deep features as a perceptual metric.
\newblock In \emph{CVPR}, 2018.

\bibitem[Zhao et~al.(2026{\natexlab{a}})Zhao, Zhu, Zheng, Zhou, Yan, Li, Yang, Li, and Zhu]{zhao2026causal}
Min Zhao, Hongzhou Zhu, Kaiwen Zheng, Zihan Zhou, Bokai Yan, Xinyuan Li, Xiao Yang, Chongxuan Li, and Jun Zhu.
\newblock {Causal Forcing++}: Scalable few-step autoregressive diffusion distillation for real-time interactive video generation.
\newblock \emph{arXiv:2605.15141}, 2026{\natexlab{a}}.

\bibitem[Zhao et~al.(2025)Zhao, Fang, Huang, Wan, Soedarmadji, Liu, Li, Lin, Dai, Yan, Yang, Ning, and Wang]{zhao2024viditq}
Tianchen Zhao, Tongcheng Fang, Haofeng Huang, Rui Wan, Widyadewi Soedarmadji, Enshu Liu, Shiyao Li, Zinan Lin, Guohao Dai, Shengen Yan, Huazhong Yang, Xuefei Ning, and Yu~Wang.
\newblock {ViDiT-Q}: Efficient and accurate quantization of diffusion transformers for image and video generation.
\newblock In \emph{ICLR}, 2025.

\bibitem[Zhao et~al.(2026{\natexlab{b}})Zhao, Pan, He, Yu, Chen, Ye, Liu, Xie, and Han]{zhao2026sana}
Yuyang Zhao, Yicheng Pan, Qiyuan He, Jincheng Yu, Junsong Chen, Tian Ye, Haozhe Liu, Enze Xie, and Song Han.
\newblock {SANA-Streaming}: Real-time streaming video editing with hybrid diffusion transformer.
\newblock \emph{arXiv:2605.30409}, 2026{\natexlab{b}}.

\bibitem[Zhu et~al.(2026)Zhu, Zhao, He, Su, Li, and Zhu]{zhu2026causal}
Hongzhou Zhu, Min Zhao, Guande He, Hang Su, Chongxuan Li, and Jun Zhu.
\newblock {Causal Forcing}: Autoregressive diffusion distillation done right for high-quality real-time interactive video generation.
\newblock In \emph{ICML}, 2026.

\end{thebibliography}
